%% file: main.tex
\documentclass[letterpaper, 10 pt, conference]{ieeeconf}  

\IEEEoverridecommandlockouts                              

\usepackage[accsupp]{axessibility}  

\usepackage{times}
\usepackage{url}

\usepackage{epsfig}
\usepackage{amssymb}
\usepackage{booktabs}
\usepackage{multirow}
\usepackage[export]{adjustbox}
\usepackage{float}
\usepackage{dashrule}
\usepackage{arydshln}
\usepackage[utf8x]{inputenc}
\usepackage[space]{grffile}
\usepackage{anyfontsize}
\usepackage{t1enc}
\usepackage{mathtools, nccmath}
\usepackage[dvipsnames]{xcolor}
\usepackage{amsmath,amsfonts}
\usepackage{algorithmic}
\usepackage{algorithm}
\usepackage{array}
\usepackage{textcomp}
\usepackage{stfloats}
\usepackage{verbatim}
\usepackage{graphicx}
\usepackage{cite}
\usepackage[table]{xcolor} 

\title{\LARGE \bf
Annotation Free Spacecraft Detection and Segmentation using Vision Language Models
}

\author{Samet Hicsonmez, Jose Sosa, Dan Pineau, Inder Pal Singh, Arunkumar Rathinam, \\ Abd El Rahman Shabayek, and Djamila Aouada%
\thanks{The authors are with Interdisciplinary Centre for Security, Reliability and Trust (SnT), University of Luxembourg, Luxembourg. Email: \texttt{\{firstname.lastname\}@uni.lu}. 
\newline
This work is supported by the Luxembourg National Research Fund (FNR), under the project reference \texttt{C21/IS/15965298/ELITE}. The experiments were performed on the Luxembourg National Supercomputer MeluXina. 
}%
}

\input{preamble}

\begin{document}

\maketitle
\thispagestyle{empty}
\pagestyle{empty}

\input{sec/0_abstract}    
\input{sec/1_introduction}
\input{sec/2_related_work}
\input{sec/3_method}

\input{sec/4_experiments}

\input{sec/5_conclusion}

{
    \small
    \bibliographystyle{IEEETran}
    \bibliography{main}
}

\end{document}

%% file: preamble.tex
\newlength \mywidth
\newlength \myw
\definecolor{myblue}{HTML}{006EAF}
\definecolor{myred}{HTML}{CC0000}

\newcommand{\VFM}{{VLM}}
\newcommand{\vfm}{{Vision Language Model}}

\usepackage{tikz}
\usetikzlibrary{shapes.geometric, positioning, fit, arrows.meta, calc, backgrounds}

\definecolor{vlmblue}{RGB}{197, 224, 243}
\definecolor{studentred}{RGB}{247, 187, 179}
\definecolor{studentblue}{RGB}{197, 224, 243}
\definecolor{augteal}{RGB}{204, 238, 239}
\definecolor{bglabel}{RGB}{240, 255, 240}
\definecolor{bgrefine}{RGB}{255, 245, 230}
\definecolor{bgdistill}{RGB}{240, 240, 255}
\definecolor{bginference}{RGB}{245, 245, 245}

\usepackage{hyperref}
\usepackage[capitalize]{cleveref}

\Crefname{figure}{Fig.}{Figs.}
\Crefname{equation}{Eq.}{Eqs.}
\Crefname{table}{Tab.}{Tabs.}
\Crefname{section}{Sec.}{Secs.}

%% file: sec/0_abstract.tex
\begin{abstract}
 
{\vfm}s ({\VFM}s) have demonstrated remarkable performance in open-world zero-shot visual recognition. However, their potential in space-related applications remains largely unexplored. In the space domain, accurate manual annotation is particularly challenging due to factors such as low visibility, illumination variations, and object blending with planetary backgrounds. Developing methods that can detect and segment spacecraft and orbital targets without requiring extensive manual labeling is therefore of critical importance. In this work, we propose an annotation-free detection and segmentation pipeline for space targets using {\VFM}s. Our approach begins by automatically generating pseudo-labels for a small subset of unlabeled real data with a pre-trained {\VFM}. These pseudo-labels are then leveraged in a teacher–student label distillation framework to train lightweight models. Despite the inherent noise in the pseudo-labels, the distillation process leads to substantial performance gains over direct zero-shot {\VFM} inference. Experimental evaluations on the SPARK-2024, SPEED+, and TANGO datasets on segmentation tasks demonstrate consistent improvements in average precision (AP) by up to 10 points.
Code and models are available at~\url{https://github.com/giddyyupp/annotation-free-spacecraft-segmentation}.

\end{abstract}

%% file: sec/1_introduction.tex
\section{INTRODUCTION}

Reliable detection and localization of spacecraft is a crucial step in achieving safe and autonomous operations in space~\cite{dung2021spacecraft} for missions such as in-orbit servicing and active debris removal.
Current object detection and segmentation models~\cite{murray2021mask, viggh2023training} for space targets rely on large amounts of annotated data to achieve high performance. 
However, in the space domain, collecting labeled data is highly challenging due to limited accessibility, high operational costs, and the difficulty of manual annotation under space conditions (see Figure~\ref{fig:teaser}). To address the scarcity of labeled data, synthetic datasets~\cite{speed, spark} generated in simulation environments~\cite{olivares2023zero} are widely used as the primary training source. Such datasets are scalable and straightforward to generate and annotate. However, models trained exclusively on synthetic data often show a significant drop in performance when applied to real-world scenarios~\cite{sda_1} due to the domain gap.

{\VFM}s have shown unprecedented performance in open-world zero-shot visual recognition tasks such as classification~\cite{radford2021learning}, object detection~\cite{liu2024grounding}, and semantic segmentation~\cite{ren2024grounded, kirillov2023segment, zhang2023simple, zou2023segment}. However, for particular domains, 
their zero-shot performance often falls short compared to models trained or fine-tuned with annotated data~\cite{feng2025vision}.
In the space domain, only a few studies have explored {\VFM}s~\cite{ulmer2025important, zhang2024automatic} for spacecraft detection or segmentation. Nevertheless, these methods do not fully exploit the zero-shot capabilities of {\VFM}s. Instead, they use them mainly as auxiliary tools for supervised training, providing limited analysis, and presenting sub-par zero-shot performances.

\input{figures/teaser}

Beyond the space domain, several strategies have been proposed to improve model performance without requiring additional manual annotations. For instance, Test-Time Augmentation (TTA)~\cite{krizhevsky2012imagenet, resnet, kimura2021understanding, vu2024test, samet2025losc} is a widely used method to improve model performance during inference by aggregating predictions coming from different views of the input image. For object detection, the predictions are usually aggregated using Non-Maximum Suppression (NMS) variants~\cite{neubeck2006efficient, bodla2017soft} with a predefined overlap ratio. However, this approach eliminates many good candidates and considers only the most confident prediction. Recently, Weighted Box Fusion (WBF)~\cite{shanmugam2021better} was proposed to adjust the final prediction based on the confidence of the overlapped detections. Another line of research explores Knowledge Distillation (KD)~\cite{hinton2015distilling} techniques, which transfer the knowledge of large pre-trained teacher models into smaller and more efficient student models~\cite{sohn2020simple, xu2021end, touvron2021training, xie2020self, sanh2019distilbert, jiao2020tinybert, wu2023tinyclip}. Usually, the distilled small student model achieves either on-par or better performance compared to the larger teacher model, thanks to the regularization effect of pseudo-labels and reduced overfitting of the small model size.

Building on these ideas, we propose an end-to-end annotation-free framework to improve the zero-shot performance of {\VFM}s for spacecraft detection and segmentation. Our method begins with a small set of unlabeled real-world images, for which pseudo-labels are automatically generated using a pre-trained {\VFM}. To enhance their reliability, we refine these pseudo-labels through TTA and WBF. The refined pseudo-labels are then employed in a teacher–student distillation scheme, where compact student models are trained using the pseudo-labeled data. This process not only boosts performance over direct zero-shot inference, but also yields lightweight models that are efficient and practical for real-world deployment.

We conduct extensive experiments on three large-scale space datasets and validate that our approach significantly improves the initial zero-shot detection and segmentation performance. 
To the best of our knowledge, the proposed pipeline is the first to exploit {\VFM}s for the challenging task of spacecraft detection and segmentation. This approach bypasses the need for extensive manual annotation, a major bottleneck in traditional computer vision methods for space applications. By leveraging {\VFM}s, our pipeline can automatically create high-quality pseudo-labels from a small, unlabeled dataset, drastically reducing the time and cost associated with data preparation.
The proposed method offers a scalable and adaptable solution for a wide range of computer vision tasks in the aerospace domain and beyond. By eliminating the reliance on manually labeled datasets, it enables rapid development and deployment of robust object detection and segmentation models for new and complex Space Situational Awareness (SSA) scenarios, such as tracking and monitoring space debris motion.

%% file: figures/teaser.tex
\renewcommand{\arraystretch}{3}

\begin{figure}[t]
\centering
\setlength\tabcolsep{1pt} 
\resizebox{1.0\columnwidth}{!}{
\begin{tabular}{cccc}

{\textit{\textbf{Input}}} & {\textit{\textbf{Our Method}}} & {\textit{\textbf{Input}}} & {\textit{\textbf{Our Method}}}   \\

\includegraphics[width=0.16\textwidth,  ,valign=m, keepaspectratio,] {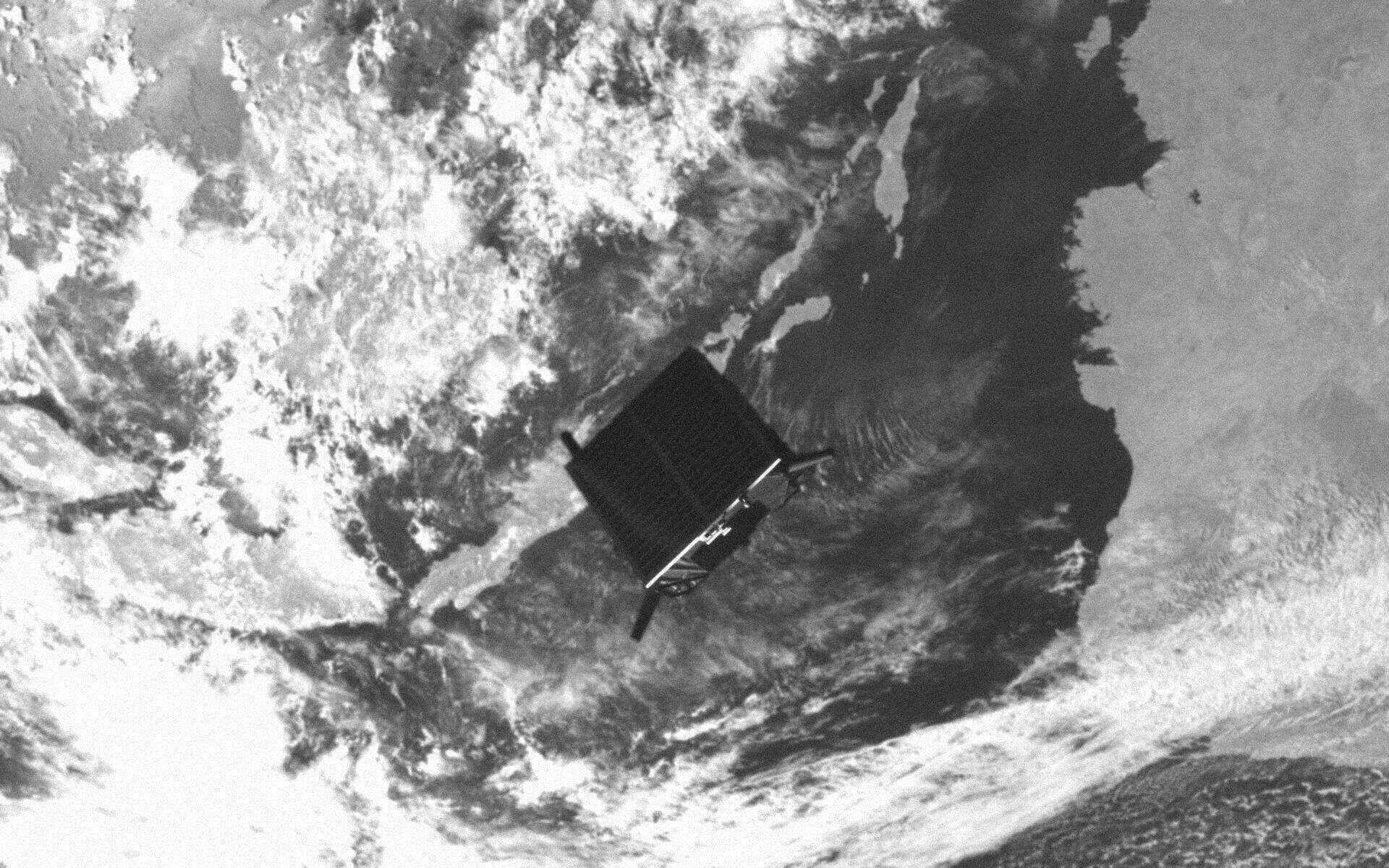} &
\includegraphics[width=0.16\textwidth,  ,valign=m, keepaspectratio,] {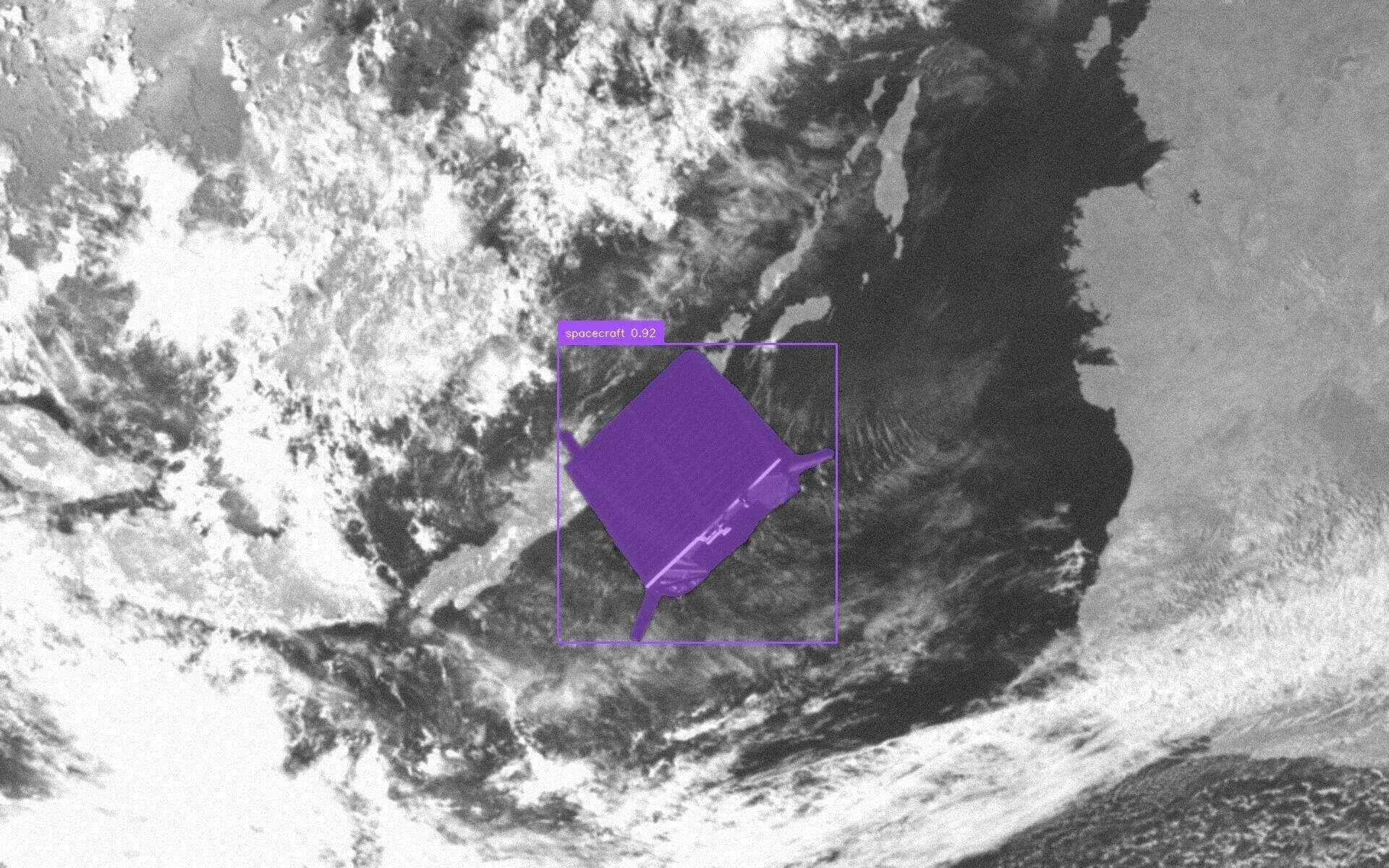} &
\includegraphics[width=0.16\textwidth,  ,valign=m, keepaspectratio,] {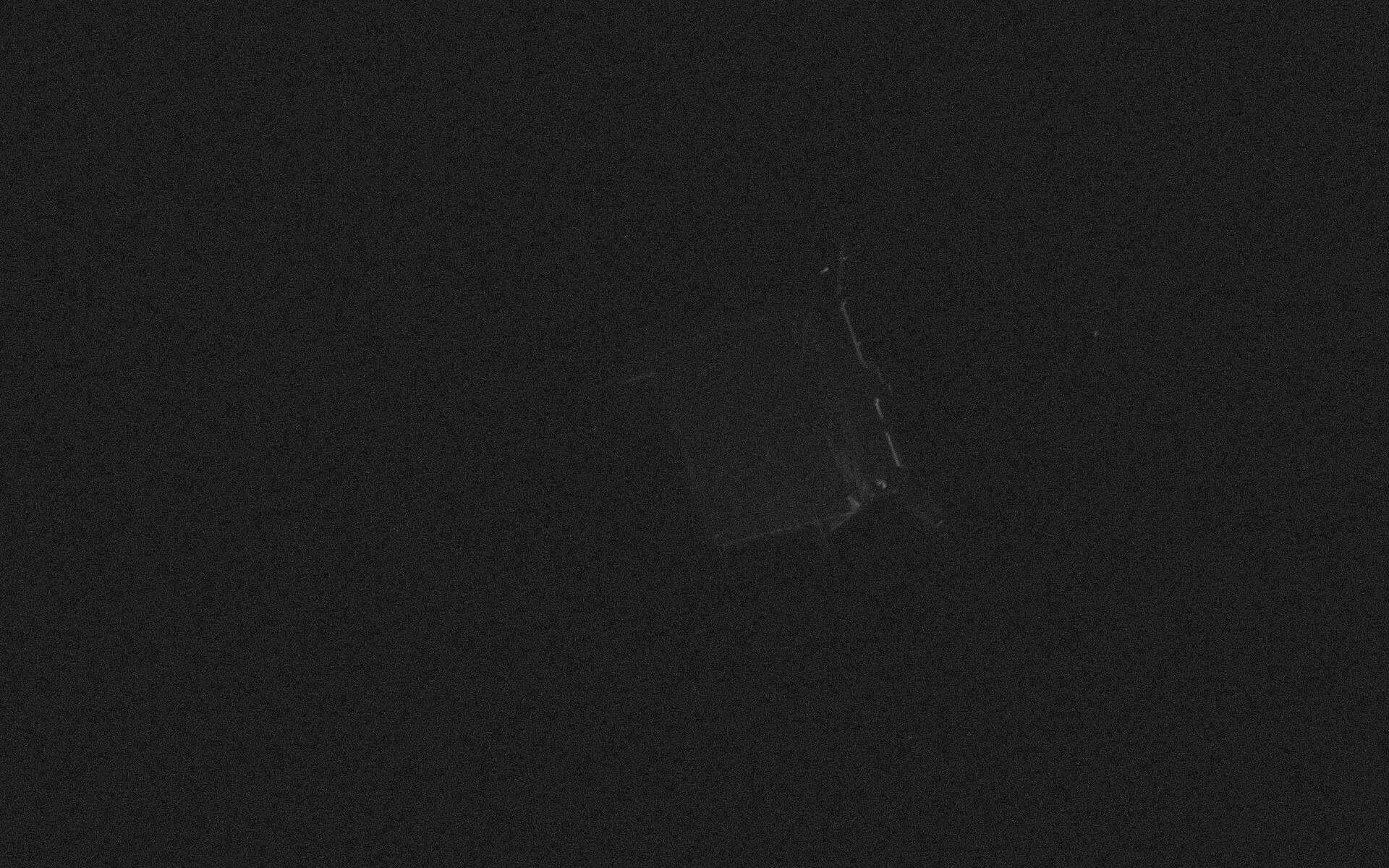} &
\includegraphics[width=0.16\textwidth,  ,valign=m, keepaspectratio,] {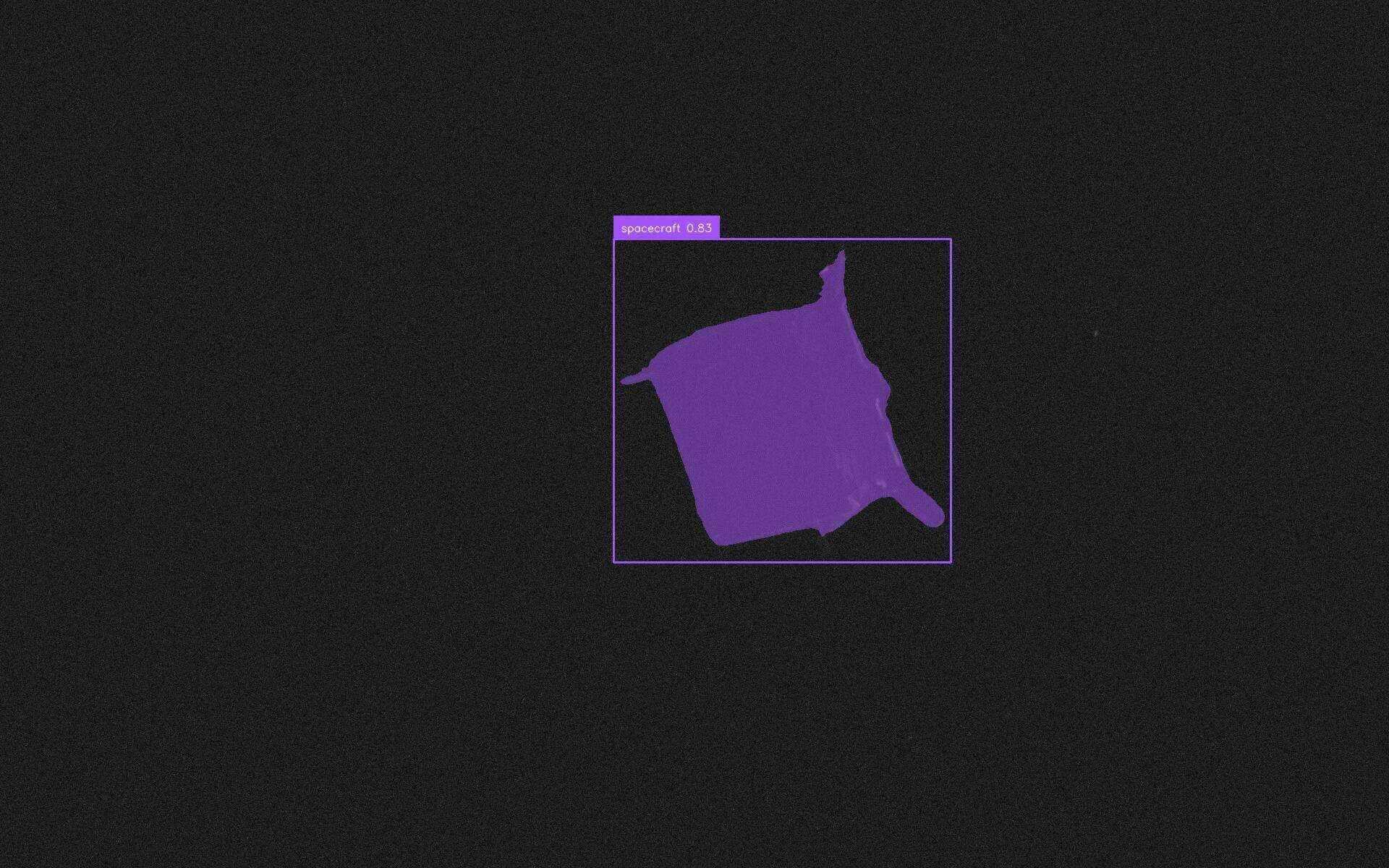}  \\

\includegraphics[width=0.16\textwidth,  ,valign=m, keepaspectratio,] {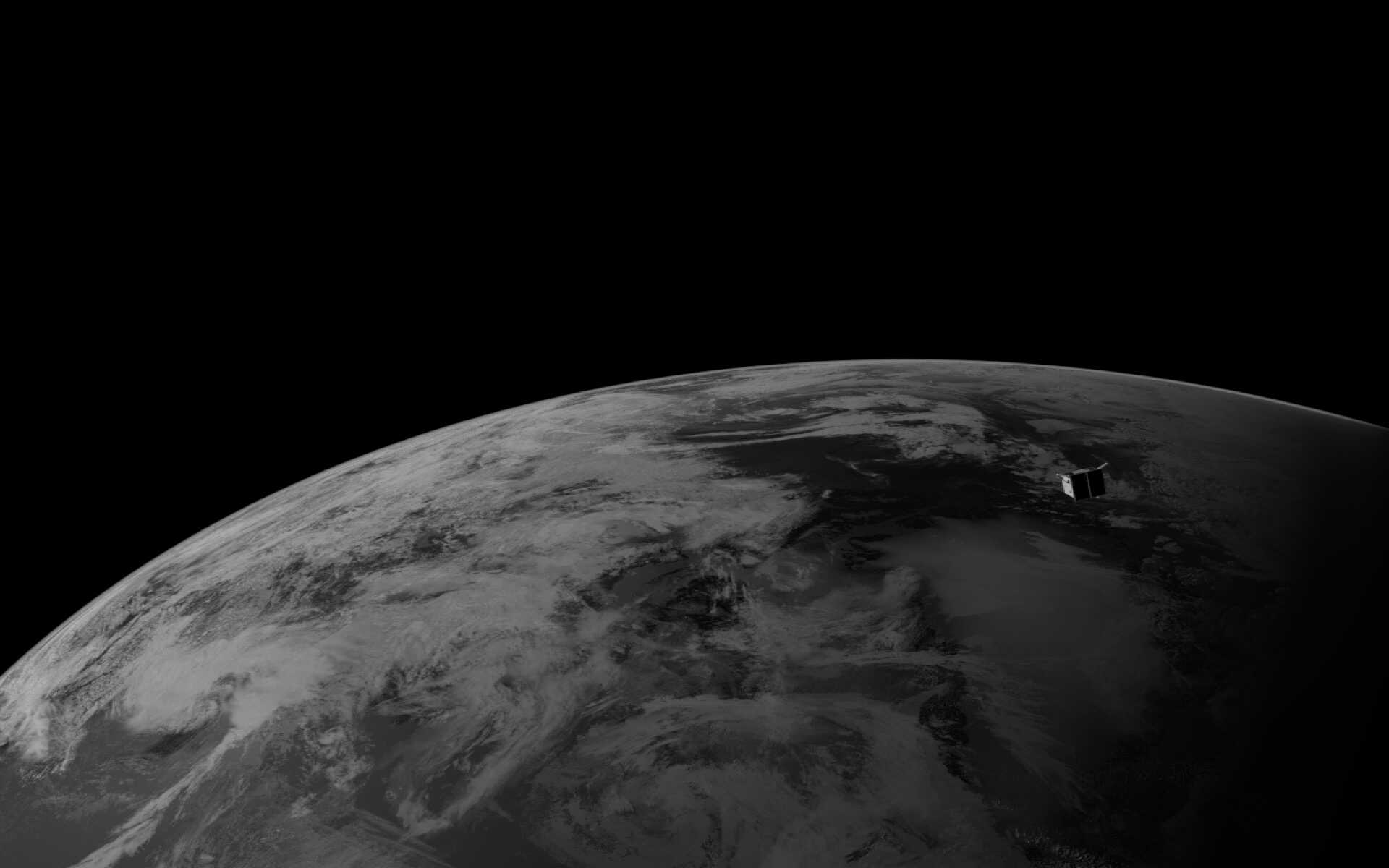} &
\includegraphics[width=0.16\textwidth,  ,valign=m, keepaspectratio,] {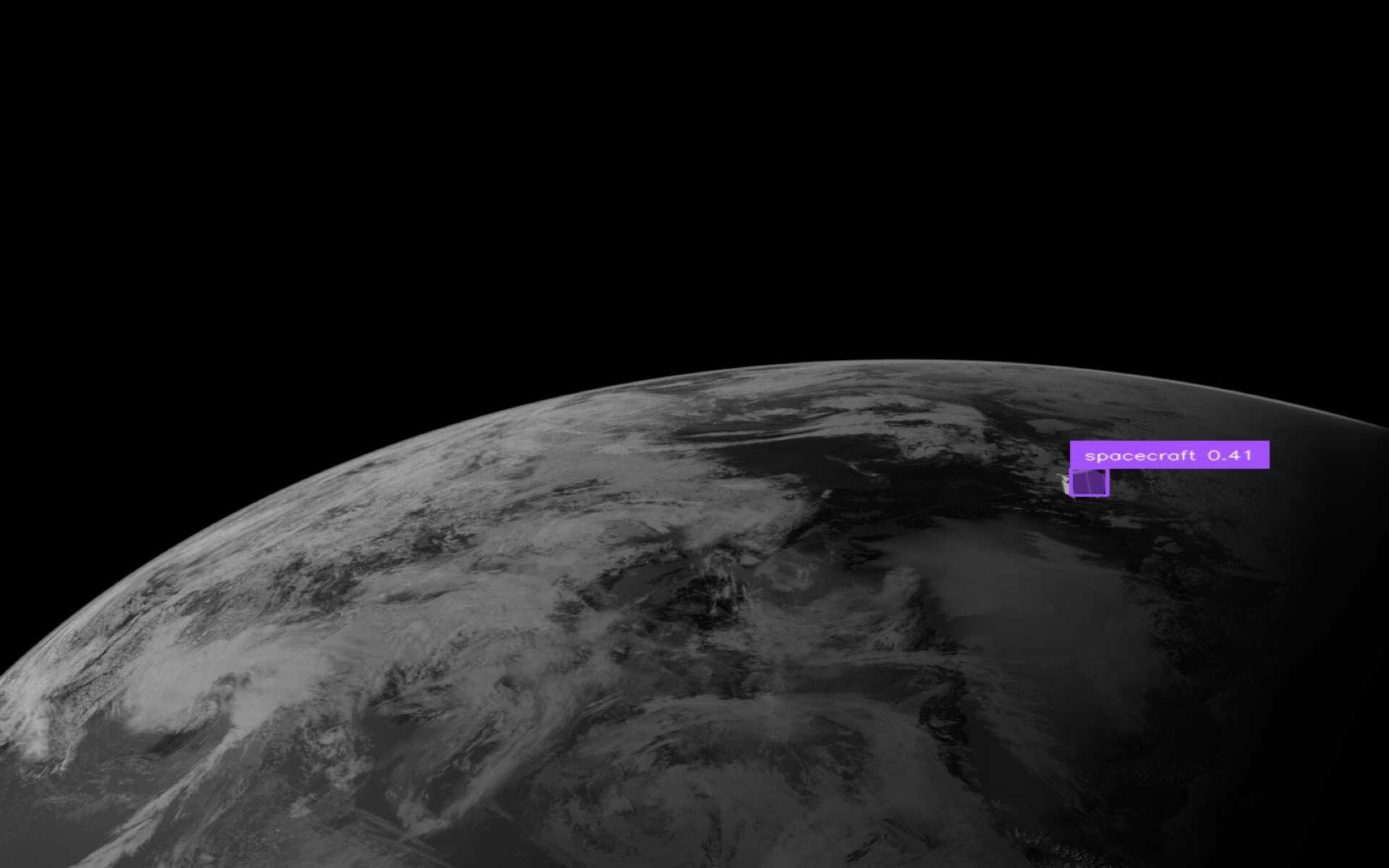} &
\includegraphics[width=0.16\textwidth,  ,valign=m, keepaspectratio,] {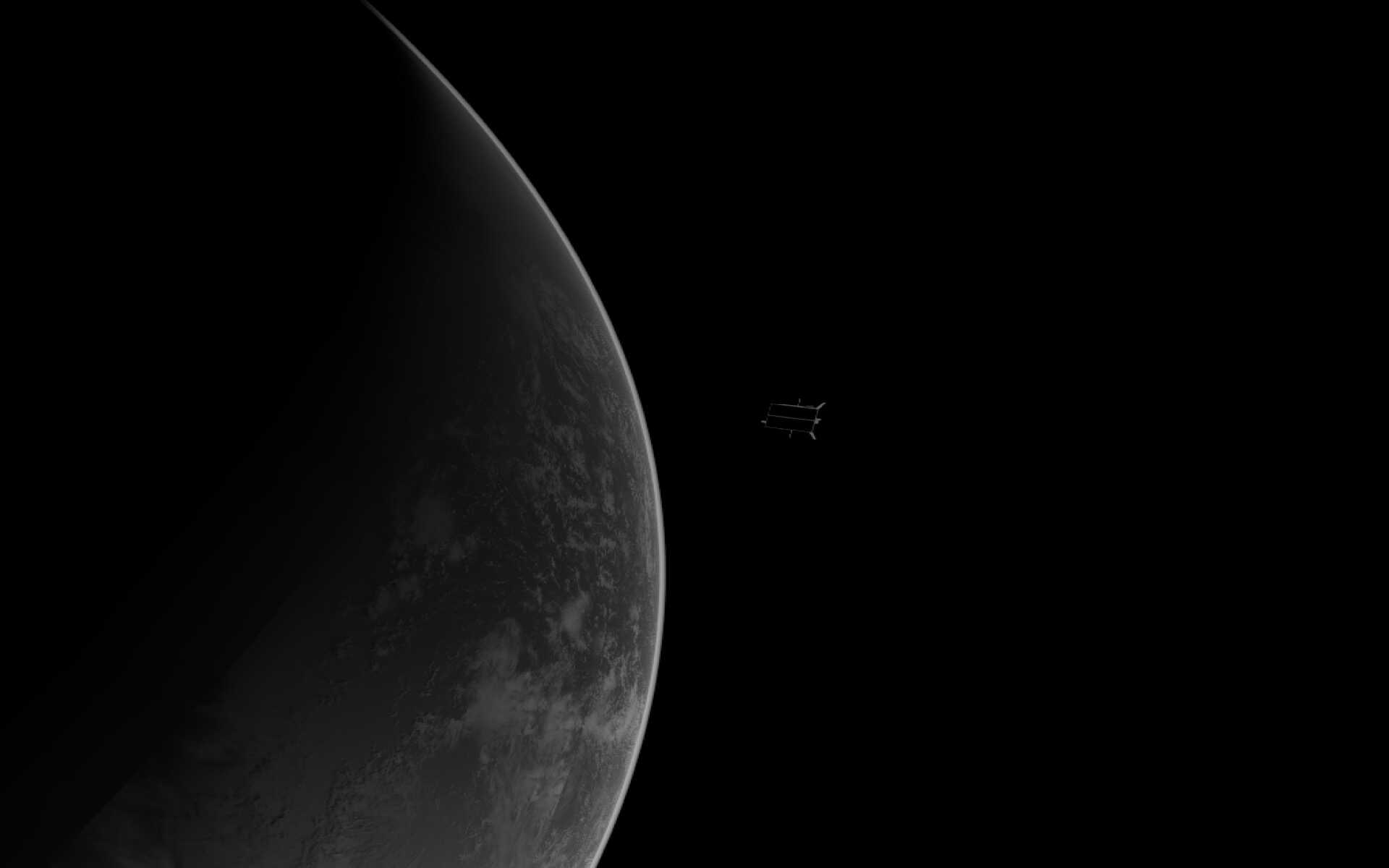} &
\includegraphics[width=0.16\textwidth,  ,valign=m, keepaspectratio,] {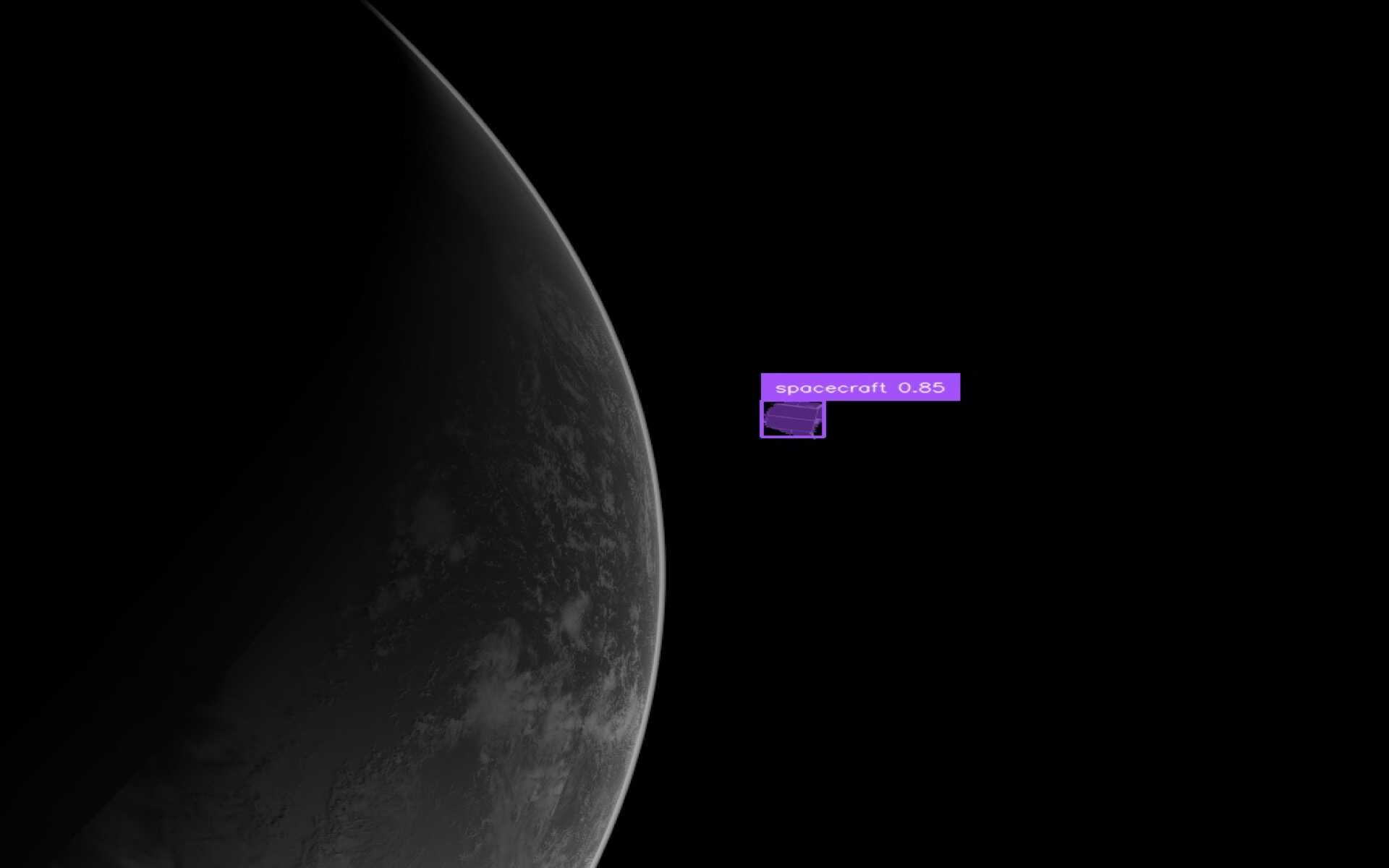}  \\

\end{tabular}}
\caption{We present challenging examples taken from SPEED+~\cite{speed} (top) and TANGO~\cite{tango} (bottom) where annotation is difficult due to the cluttered background and low visibility. Our method detects and segments the spacecraft object with high precision in all cases. Zoom in for details.}
\label{fig:teaser}
\vspace{-2mm}

\end{figure}
\renewcommand{\arraystretch}{1}

%% file: sec/2_related_work.tex
\section{RELATED WORK}

This section presents the relevant works on {\vfm}s and their applications in the space domain, followed by TTA methods on improving inference robustness and accuracy. Finally, recent developments in knowledge distillation to achieve an efficient student model are presented.

\subsection{{\vfm}s}

Recent {\VFM}s~\cite{radford2021learning, jia2021scaling} 
show strong zero-shot capabilities in image-level tasks. Open-vocabulary detection and segmentation models~\cite{zhang2023simple, zou2023segment, ren2024grounded, liu2024grounding, ren2024grounding, li2023semantic} go further and extend this capability to dense pixel-level tasks.

OpenSeed~\cite{zhang2023simple} aligns textual embeddings from pre-trained language models with dense pixel-level features from a mask proposal backbone, enabling accurate segmentation of unseen categories. The method introduces a joint training strategy with contrastive objectives that bridge vision and language modalities, achieving competitive performance across open-vocabulary semantic and panoptic segmentation benchmarks. SEEM~\cite{zou2023segment} extends this line of work by proposing a unified model capable of handling multiple segmentation tasks, i.e. semantic, instance, panoptic, and interactive, through flexible conditioning on prompts such as text, points, or boxes. This promptable design demonstrates strong generalization and adaptability, making SEEM a versatile framework for open-world scenarios. 

Grounded-SAM~\cite{ren2024grounded}, on the other hand, integrates the grounding capabilities of GroundingDINO~\cite{liu2024grounding}, i.e. the ability to connect and align textual descriptions with specific objects or regions within an image, with the high-quality mask generation of SAM~\cite{kirillov2023segment}, enabling open-vocabulary detection and segmentation directly from natural language queries. By leveraging the complementary strengths of grounding and segmentation, Grounded-SAM achieves robust zero-shot generalization across diverse object categories. 

Despite the success of {\VFM}s in natural images, their performance on space applications has not been fully explored. Recently,~\cite{ulmer2025important} experimented with different text prompts to detect spacecraft bounding boxes in a zero-shot manner on the SPEED+~\cite{speed} dataset using GroundingDINO~\cite{liu2024grounding}. However, their work is limited to a single dataset and task, and fails to exploit the full potential of {\VFM}s (c.f. see Table~\ref{tab:distill_res}).

In~\cite{zhang2024automatic}, authors propose to use SAM~\cite{kirillov2023segment} to automatically segment spacecraft. They train an object detection model within a supervised framework. Then, they use the predicted boxes to prompt the SAM. This approach assumes the availability of the bounding boxes to first train the object detection method. However, our method does not require any labeled data and performs detection and segmentation simultaneously.

\subsection{Test-time Augmentation (TTA)}

TTA is a widely used technique in various visual detection tasks to improve robustness and accuracy at inference time~\cite{krizhevsky2012imagenet, simonyan2014very, resnet, shanmugam2021better,kim2020learning, kimura2021understanding, vu2024test, moshkov2020test}. The core idea is to apply a set of transformations such as horizontal or vertical flipping, scaling, and color perturbations to the input image, run the model on each augmented version, and then aggregate the predictions. While simple ensembling across augmentations can yield consistent gains, for object detection, it introduces the challenge of merging multiple overlapping detections. 

Traditional strategies such as Non-Maximum Suppression (NMS)~\cite{neubeck2006efficient, bodla2017soft} often discard potentially useful predictions when bounding boxes overlap, leading to suboptimal results. To address this, Weighted Boxes Fusion (WBF)~\cite{solovyev2021weighted} has emerged as a more effective merging strategy. Instead of suppressing overlapping boxes, WBF computes a weighted average of their coordinates, where the weights are determined by the associated confidence scores. This produces more accurate bounding boxes and preserves valuable information from multiple augmented predictions. Recent studies have shown that combining TTA with WBF provides consistent improvements in detection and segmentation benchmarks, striking a balance between robustness and precision.

\subsection{Label Distillation}

Knowledge distillation~\cite{hinton2015distilling} is a well-established approach for model compression, in which a high-capacity teacher network transfers its learned representations to a smaller and more efficient student model. 
The pseudo labels coming from the teacher model could be treated as ground truth by directly using them as hard labels, i.e. one-hot encoded. However, this approach overlooks the possible low-confidence detections. One approach to overcome this issue is using soft labels~\cite{li2017learning, yuan2020revisiting, zi2021revisiting, yang2023knowledge}, i.e. incorporating the prediction confidences with specialized loss functions. Generalized Focal Loss~\cite{li2020generalized} extends focal loss by jointly modeling classification confidence and localization quality, while Varifocal Loss~\cite{zhang2021varifocalnet} builds on this idea by weighting high-quality samples more heavily, enabling detectors to better exploit predictions aligned with their IoU scores.

An alternative approach to reduce the pseudo label noise is applying confidence thresholding~\cite{lee2013pseudo, sohn2020simple, liuunbiased} to the detections. In this setting, only the high-quality detections of the teacher are used to train the student model. Due to its simplicity, we prefer to use this approach in our trainings.

%% file: sec/3_method.tex
\section{METHOD}
\label{sec:method}

\begin{figure*}[t]
\centering
\resizebox{1.0\textwidth}{!}{
\input{figures/model_pipeline}
}
\vspace{-3mm}
\caption{The processing pipeline of our method. Given unlabeled training images, we first automatically annotate them using a pretrained VLM with a fixed prompt of ``spacecraft". In the next stage, we refine these pseudo-annotated images leveraging test-time augmentations. The refined labels are distilled to a shallow Student Model which is employed for the inference.}
\label{fig:model}
\end{figure*}
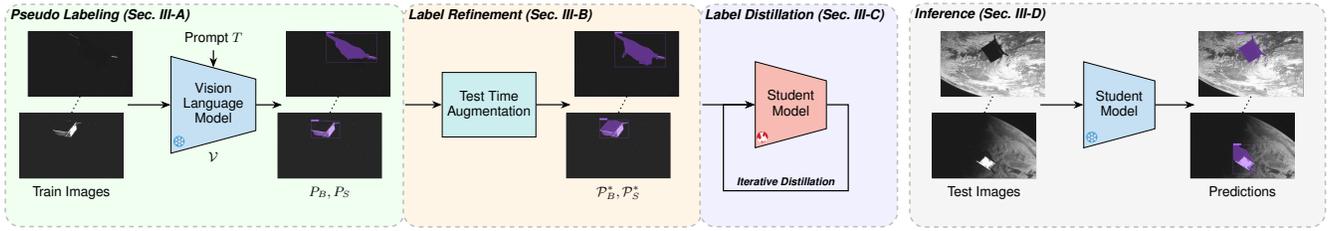

We propose a pipeline for spacecraft detection and segmentation that does not assume the availability of manually annotated data. Our approach exploits the zero-shot capabilities of pre-trained {\VFM}s to generate pseudo-labels, which are subsequently refined through an iterative teacher–student distillation framework. Then, the student model is deployed for inference. Figure~\ref{fig:model} illustrates the overall workflow, which comprises four stages:
\begin{enumerate}
    \item \textbf{Pseudo-labeling} stage utilizes pre-trained {\VFM}s to generate necessary pseudo-labels. 
    \item \textbf{Label refinement} stage adopts TTA together with WBF and confidence-based thresholding to remove noisy or inaccurate predictions.
    \item \textbf{Label distillation} stage implements an iterative distillation process involving a teacher and student model. The teacher produces pseudo-labels to train a simpler student model.
    \item \textbf{Inference} stage employs solely the student model that has been trained with refined pseudo-labels.
\end{enumerate}

The following subsections will describe in more detail each of the steps in our proposed pipeline.

\subsection{Pseudo-label generation}
\label{sec:pseudo}

The first stage of our pipeline employs a pre-trained {\VFM}s to automatically annotate a given set of images. Rather than relying on manual labeling, the {\VFM}s are prompted to produce pseudo-labels that capture instance segmentation masks and bounding box predictions.

Formally, for a given image $I$ and a textual prompt $T$, the {\VFM} $\mathcal{V}$ produces 2D instance segmentation and bounding box predictions $P_S, P_B = \mathcal{V}(I, T)$ respectively. 
Since the space datasets contain a single instance in images, we choose only the top detection and/or segmentation predictions of the models. 
This step generates a single pseudo- box and mask annotation for each training image.

To be concise, the subsequent sub-sections focus on the particulars of bounding box predictions. However, this approach is equally applicable to segmentation masks.

\subsection{Label refinement}
\label{sec:refinement}

Given the domain gap between the data used to pre-train {\VFM}s and the nature of space data, their zero-shot predictions might be inaccurate. To address this limitation, we introduce a refinement stage within our pipeline, where the initially generated pseudo-labels are improved to mitigate noise and enhance label quality. In particular, we follow standard refinement strategies, such as using TTA with WBF and confidence-based thresholding.

\subsubsection{Refinement using TTA}

We use test time augmentation strategy to enhance the quality of pseudo-labels by first compiling a set of augmentations applied to input images, followed by pooling the predictions using WBF. 

Let $\mathcal{A} = \{ a_1, a_2, \dots, a_K \}$ represent the formally defined augmentation set. For an input image $I$, $K$ augmented versions are generated, each determined by an augmentation in $\mathcal{A}$:
\begin{equation}
 I_k = a_k(I), \quad k = 1, \dots, K.
\end{equation}

Then, each augmented image is passed through $\mathcal{V}(\cdot)$, which predicts a set of bounding boxes:
\begin{equation}
\mathcal{P}_{B_k} = \mathcal{V}(I_k, T) = \{ (b_i^{(k)}, s_i^{(k)}, \ell_i^{(k)}) \}_{i=1}^{N_k},
\end{equation}

where $b_i^{(k)} \in [0,1]^4$ are the normalized box coordinates, $s_i^{(k)} \in [0,1]$ is the confidence score, and $\ell_i^{(k)}$ is the predicted class. Each box is then mapped back to the original image coordinates using the inverse transform $a_k^{-1}$:
\begin{equation}
\tilde{b}_i^{(k)} = a_k^{-1}\!\left(b_i^{(k)}\right).
\end{equation}

Let $\mathcal{B} = \{ \tilde{b}_i^{(k)}, s_i^{(k)}, \ell_i^{(k)} \}$ be the union of all predictions from all the augmentations.
WBF groups boxes with Intersection-over-Union (IoU) greater than a threshold $\tau$.

For each cluster $C = \{ (b_j, s_j) \}_{j=1}^M$, the fused box $b^\ast$ is computed as a weighted average:
\begin{equation}
\begin{aligned}
b^\ast &= \frac{\sum_{j=1}^M s_j \, b_j}{\sum_{j=1}^M s_j} ,
&\qquad
s^\ast &= \frac{1}{M} \sum_{j=1}^M s_j
\end{aligned}
\end{equation}

where $s^\ast$ is the fused confidence score. The assigned class label is determined by majority voting or weighted voting based on scores.

The final set of predictions after TTA and WBF is defined:
\begin{equation}
\mathcal{P}_B^\ast = \{ (b^\ast_m, s^\ast_m, \ell^\ast_m) \}_{m=1}^{M^\ast}.
\end{equation}

\subsubsection{Refinement using confidence based thresholding}

Upon implementing WBF, we acquire a set of fused predictions with associated confidence scores. However, when aggregated predictions show low IoU, WBF often leads to a significant reduction in confidence. To further enhance the quality of the predictions, we applied a confidence-based thresholding. This filtering procedure removes noisy, low-confidence predictions while retaining only those with high reliability.
\begin{equation}
\mathcal{P}_B{^{\ast}_{\theta}} 
= \Big\{ (b_m^{\ast},\, s_m^{\ast},\, \ell_m^{\ast}) 
\;\big|\; s_m^{\ast} \geq \theta \Big\},
\end{equation}
where $\theta \in [0,1]$ is the confidence threshold.

\subsection{Label distillation}
\label{sec:distillation}

In this stage, the refined pseudo-labels are considered as outputs from a teacher model, and we train a shallow student model using iterative knowledge distillation. This phase offers three key advantages: i) label-distilled student models often outperform the teacher models, ii) a significant reduction in model size is achieved, and iii) the more compact model can operate in real-time and facilitate autonomous operations.

Given the confidence-filtered TTA pseudo box predictions $\mathcal{P_B}^{\ast}_{\theta}$ generated by the teacher model ({\VFM}), the predictions $\mathcal{Q}$ from the student model $S_\phi$ for a specific image $I$ are defined as:
\begin{equation}
\mathcal{Q} = S_\phi(I) = \{(b_j, p_j)\}_{j=1}^{N_S},
\end{equation}

where $b_j$ and $p_j$ are the predicted box coordinates and associated class labels, respectively.  

In the distillation process, the student model undergoes training with the standard supervised loss:
\begin{equation}
\mathcal{L}_{\text{distill}}
= 
\alpha \, \mathcal{L}_{\text{cls}}\!\left(p, \hat{y} \right)
+ \beta \, \mathcal{L}_{\text{reg}}\!\left(b, \hat{b} \right)
\end{equation}

where $\hat{b}, \hat{y}$ are the box and class predictions of the teacher {\VFM}, respectively. $p, b$ are the student’s predicted class probability and box, and finally, $\alpha, \beta$ are the weighting factors between the classification loss $\mathcal{L}_{\text{cls}}$ and the regression loss $\mathcal{L}_{\text{reg}}$.

Upon training the student model, its predictions are employed to assign new labels to the training images, acting as pseudo-labels. These are used to retrain the student model from the ground up. This process, referred to as iterative knowledge distillation, serves to improve pseudo-label accuracy and enhance the student model's performance. Our approach employs a single iteration of this distillation process.

\subsection{Inference}
\label{sec:inference}

We utilized the distilled student model for inference. Owing to its lightweight design, the student facilitates real-time operation with minimal computational and memory requirements, ideal for in-orbit platform deployment. At the same time, it preserves much of the accuracy of the teacher model, offering a favorable trade-off between efficiency and performance.

%% file: figures/model_pipeline.tex
\begin{tikzpicture}[
    font=\sffamily\small,
    >=Stealth,
    image/.style={draw, fill=black!80, minimum width=2.4cm, minimum height=1.5cm, inner sep=0},
    model/.style={trapezium, trapezium stretches body, trapezium angle=75, draw, inner sep=2pt, minimum width=2cm, minimum height=1.5cm},
    block/.style={draw, fill=white, minimum width=1.5cm, minimum height=1cm, align=center},
    container/.style={draw=gray!50, dashed, rounded corners=10pt, line width=1pt}
]

    \begin{scope}[local bounding box=pseudo]
        \node[image] (im1) at (0,-0.1) {\includegraphics[width=2.4cm]{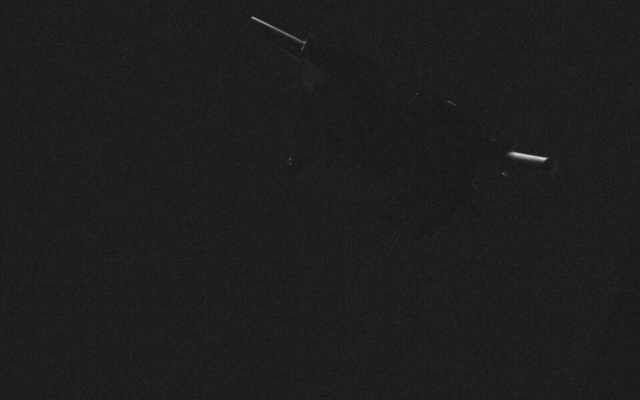}};
        \node[image] (im2) at (-0.2,-2) {\includegraphics[width=2.4cm]{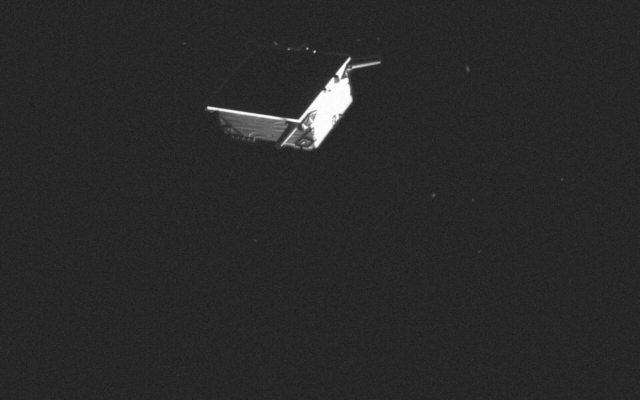}};
        \draw[dotted, thick] (im1.south) -- (im2.north);
        \node[below=2pt of im2] {{Train Images}};

        \path ($(im1.east)!0.5!(im2.east)$) coordinate (img_mid);

        \node[model, fill=vlmblue, rotate=-90, right=1cm of img_mid, anchor=south] (vlm) 
        {\rotatebox{90}{\begin{tabular}{c} Vision \\ Language \\ Model \end{tabular}}};
        \node[below=0.1cm of vlm.east] {$\mathcal{V}$};
        
        \begin{scope}
            \clip (2.3,-1.9) circle (0.13cm); 
    
            \node (sfi1) at (2.3,-1.9) {\includegraphics[width=0.3cm]{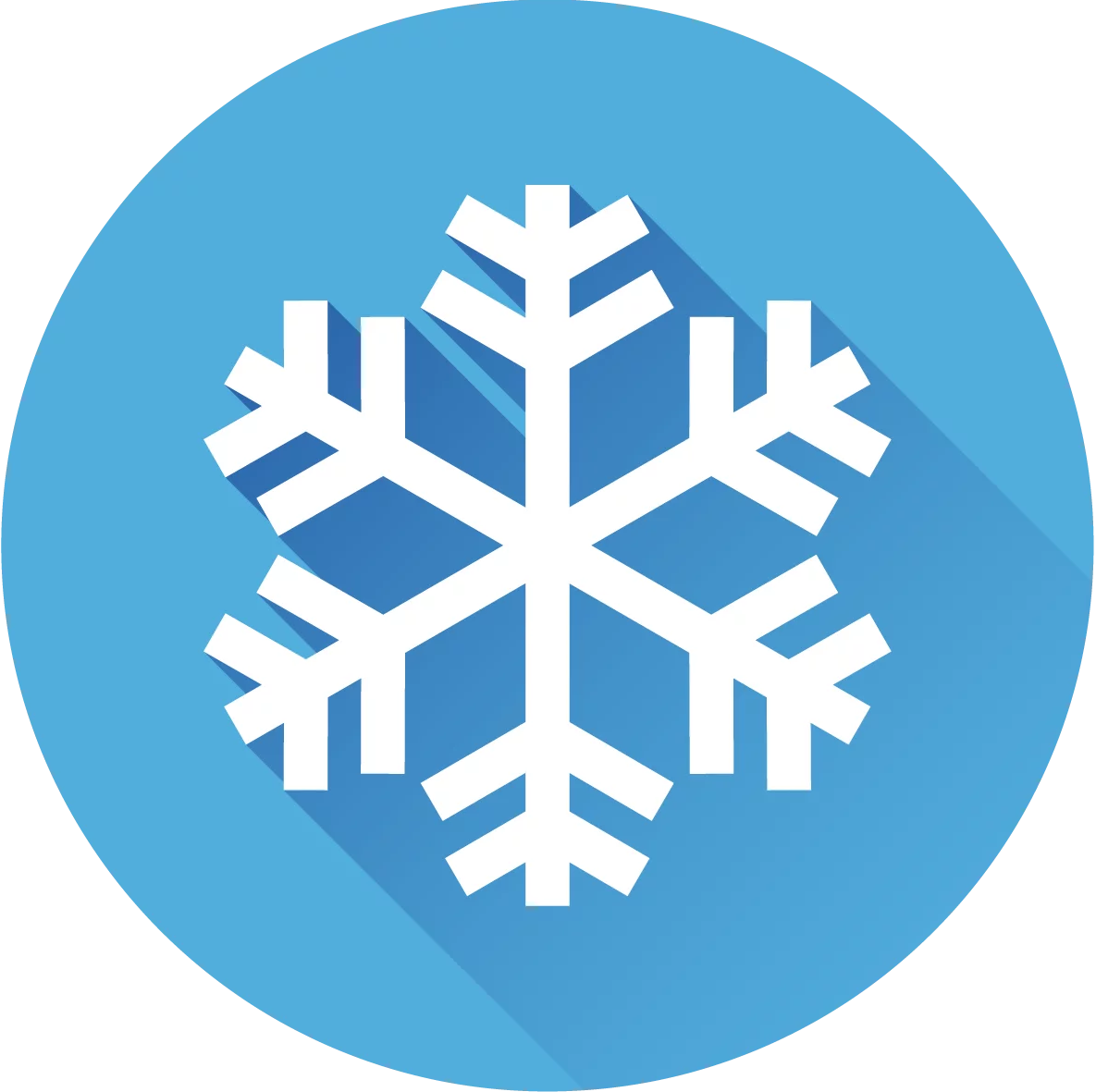}};
        \end{scope}

        \node[above=0.4cm of vlm.west] (prompt) {Prompt $T$};
        \draw[->, thick] (prompt) -- (vlm.west);

        \node[image] (out1) at (6,-0.1) {\includegraphics[width=2.4cm]{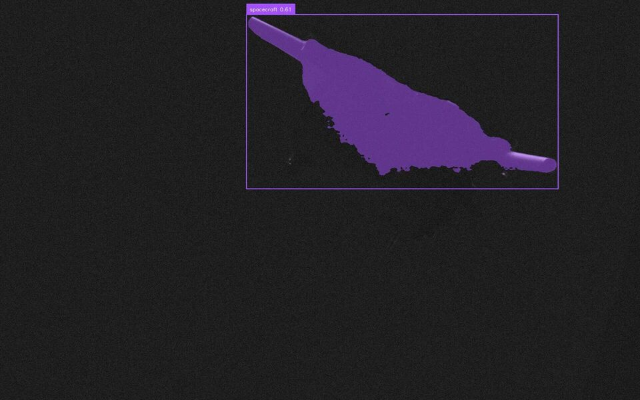}};
        \node[image] (out2) at (5.8,-2) {\includegraphics[width=2.4cm]{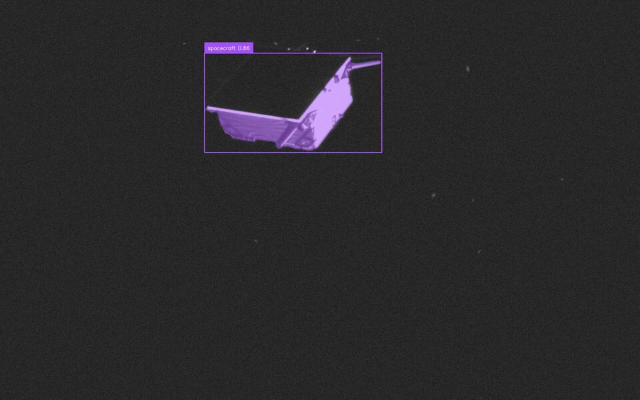}};
        \draw[dotted, thick] (out1.south) -- (out2.north);
        \node[below=2pt of out2] (out_text1) {$P_B, P_S$};

        \path ($(out1.east)!0.5!(out2.east)$) coordinate (out_mid);

        \draw[->, thick] ($(im1.east)!0.5!(im2.east)$) -- (vlm.south);
        \draw[->, thick] (vlm.north) -- ($(out1.west)!0.5!(out2.west)$);
    \end{scope}
    
    \begin{scope}[on background layer]
        \node[container, fill=bglabel, fit=(im1) (out2) (prompt) (out_text1), inner sep=15pt] (pseudo_box) {};
        \node[anchor=north west] at (pseudo_box.north west) {\textit{\textbf{Pseudo Labeling (\Cref{sec:pseudo})}}};
    \end{scope}

    \begin{scope}[local bounding box=refinement]
        \node[block, fill=augteal, right=1.95cm of pseudo_box.east, yshift=0.25cm, anchor=center, minimum height=1.5cm] (tta) 
            {Test Time\\Augmentation};
        
        \node[image, right=0.9cm of tta, yshift=0.95cm] (ref1) {\includegraphics[width=2.4cm]{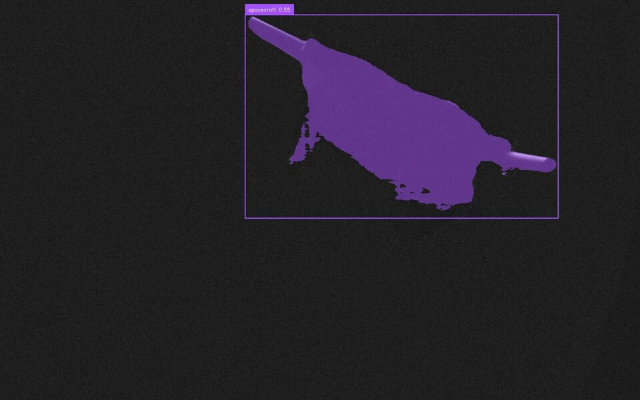}};
        \node[image, right=0.7cm of tta, yshift=-0.95cm] (ref2) {\includegraphics[width=2.4cm]{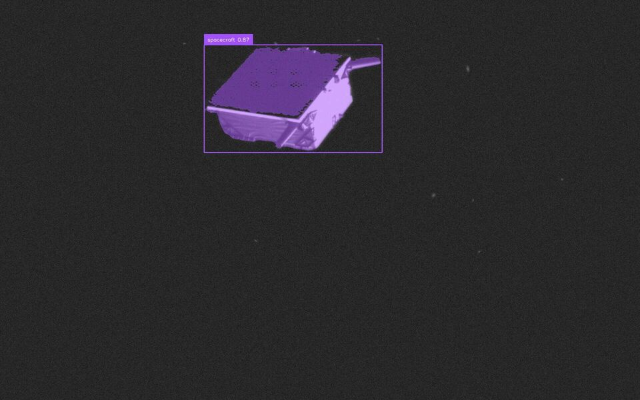}};
        \path ($(ref1.east)!0.5!(ref2.east)$) coordinate (rimg_mid);
        
        \draw[dotted, thick] (ref1.south) -- (ref2.north);

        \node[above=0.80cm of tta.north] (prompt_empty) {};
        \node[left=0.12cm of tta.west] (prompt_empty2) {};
        
        \node[below=1.8pt of ref2] (ref_label) {$\mathcal{P}_B^*, \mathcal{P}_S^*$};

        \draw[->, thick] (pseudo_box.east |- tta.west) -- (tta.west);
        
        \draw[->, thick] (tta.east) -- ($(ref1.west)!0.5!(ref2.west)$);
    \end{scope}

    \begin{scope}[on background layer]
        \node[container, fill=bgrefine, 
        fit=(tta) (ref1) (ref_label) (prompt_empty) (prompt_empty2), 
        inner sep=15pt] (refine_box) {};
        \node[anchor=north west] at (refine_box.north west) {\textit{\textbf{Label Refinement (\Cref{sec:refinement})}}};
    \end{scope}

    \begin{scope}[local bounding box=distill]

        \node[model, fill=studentred, rotate=-90, right=2.7cm of rimg_mid, anchor=center] (student_dist) {\rotatebox{90}{\begin{tabular}{c} Student \\ Model \end{tabular}}};

        \node[above=0.77cm of student_dist.west] (prompt_empty) {};
        \node[left=0.50cm of student_dist.south] (prompt_empty2) {};
        \node[right=1.70cm of student_dist.east] (prompt_empty4) {};
        \node[below=2.06cm of student_dist.north] (prompt_empty3) {};

        \begin{scope}
            \clip (15.85,-1.8) circle (0.13cm); 
    
            \node (fi1) at (15.85,-1.8) {\includegraphics[width=0.3cm]{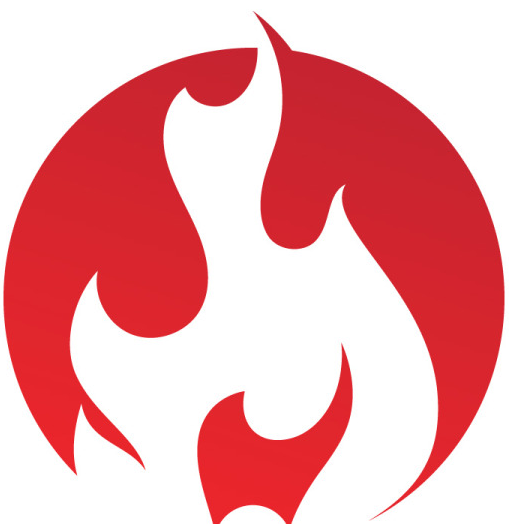}};
        \end{scope}

        \draw[->, thick] (student_dist.north) -- ++(0.5,0) -- ++(0,-2.0) -- node[above, pos=0.5] {\scriptsize\textit{\textbf{Iterative Distillation}}} ++(-2.9,0) -- ++(0,2.0) -- (student_dist.south);

        \draw[->, thick] (refine_box.east |- student_dist.south) -- (student_dist.south);
        
    \end{scope}

    \begin{scope}[on background layer]
        \node[container, fill=bgdistill, fit=(student_dist) (prompt_empty2) (prompt_empty) (prompt_empty3) (prompt_empty4), 
        inner sep=15pt] (distill_box) {};
        \node[anchor=north west] at (distill_box.north west) {\textit{\textbf{Label Distillation (\Cref{sec:distillation})}}};
    \end{scope}


    \begin{scope}[local bounding box=inference, xshift=21.2cm]
        \node[image] (tim1) at (0,-0.1) {\includegraphics[width=2.4cm]{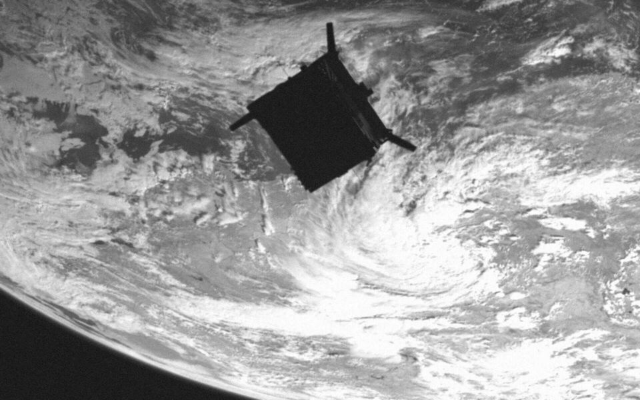}};
        \node[image] (tim2) at (-0.2,-2) {\includegraphics[width=2.4cm]{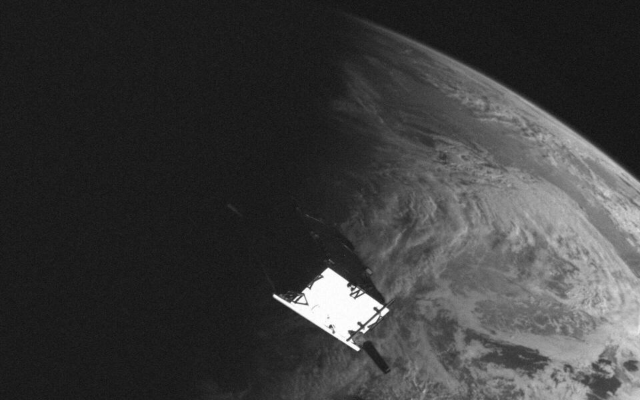}};
        \draw[dotted, thick] (tim1.south) -- (tim2.north);
        \node[below=2pt of tim2] {{Test Images}};

        \path ($(tim1.east)!0.5!(tim2.east)$) coordinate (timg_mid);

        \node[model, fill=studentblue, rotate=-90, right=1cm of timg_mid, anchor=south] (sm) 
        {\rotatebox{90}{\begin{tabular}{c} Student \\ Model \end{tabular}}};
        
        \begin{scope}
            \clip (2.3,-1.8) circle (0.13cm); 
            
            \node (sfi) at (2.3,-1.8) {\includegraphics[width=0.3cm]{images/model_fig/icon_snowflake.png}};
        \end{scope}

        \node[above=0.80cm of sm.west] (prompt) {};

        \node[image] (tout1) at (6,-0.1) {\includegraphics[width=2.4cm]{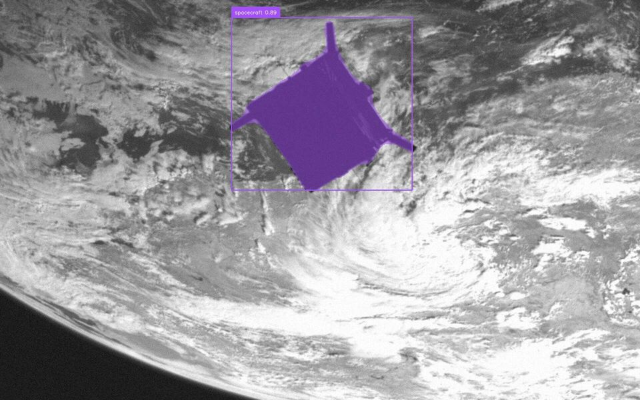}};
        \node[image] (tout2) at (5.8,-2) {\includegraphics[width=2.4cm]{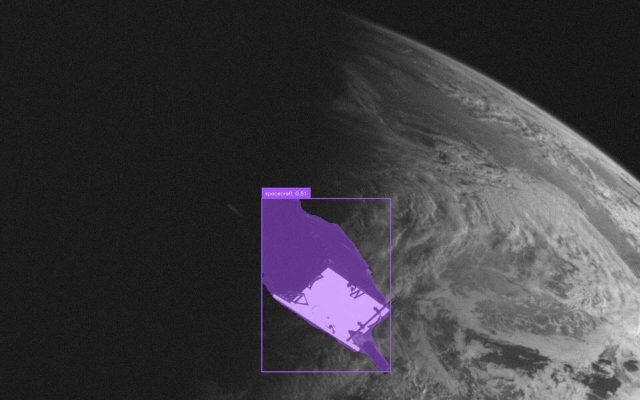}};
        \draw[dotted, thick] (tout1.south) -- (tout2.north);
        \node[below=2pt of tout2] {{Predictions}};

        \path ($(tout1.east)!0.5!(tout2.east)$) coordinate (tout_mid);

        \draw[->, thick] ($(tim1.east)!0.5!(tim2.east)$) -- (sm.south);
        \draw[->, thick] (sm.north) -- ($(tout1.west)!0.5!(tout2.west)$);
    \end{scope}
    
    \begin{scope}[on background layer]
        \node[container, fill=bginference, fit=(inference), inner sep=15pt] (inf_box) {};
        \node[anchor=north west] at (inf_box.north west) {\textit{\textbf{Inference (\Cref{sec:inference})}}};
    \end{scope}
    
\end{tikzpicture}

%% file: sec/4_experiments.tex
\section{EXPERIMENTS}

In this section, we introduce the curated datasets employed in the experiments and summarize the upper bound performances (Oracle) achieved through fully-supervised training. Subsequently, we present the zero-shot baseline performance of {\VFM}s in spacecraft detection and segmentation, followed by the impact of the label refinement and label distillation steps on enhancing this baseline.

\subsection{Datasets}
In our experiments, we use widely adopted SPARK-2024~\cite{spark}, SPEED+~\cite{speed}, and TANGO~\cite{tango} datasets. 

\textbf{SPARK-2024} dataset~\cite{spark} was part of the Spacecraft Pose Estimation Challenge at the AI4SPACE workshop held during CVPR'24. Stream-2 of the dataset is dedicated to spacecraft trajectory estimation and comprises sequences of both simulated images and real images acquired in the SnT Zero-G lab~\cite{olivares2023zero}. The training set includes $100$ groups of trajectories, each containing $300$ labeled synthetic images with ground-truth pose and bounding box annotations, but missing segmentation annotations. The test set consists of four real trajectory sequences comprising $2100$ images in total. In our experiments, we used only the test sequences and created a training split with $500$ randomly selected images and used the remaining images ($1600$) for evaluation. 

\textbf{SPEED+} dataset~\cite{speed} was part of the SPEED + challenge hosted by the European Space Agency (ESA) and provides both synthetic images and real images with high-fidelity images of the Tango spacecraft model captured under two distinct illumination conditions (Lightbox and Sunlamp). The Lightbox domain contains $6,740$ images of the spacecraft model illuminated by diffuse albedo lightboxes, which simulate Earth's orbital lighting conditions. In contrast, the Sunlamp domain consists of $2,791$ images of the same model illuminated by a metal-halide arc lamp, which approximates direct sunlight. Both splits contain box and segmentation annotations.
Normally, these subsets are only used for testing, i.e., there is no official training split. However, since our aim is to distill from large {\VFM}s, we create custom training and test datasets on these two datasets using the test splits. We follow the same procedure as SPARK-2024 and create training splits with $500$ randomly selected images and use the remaining images ($6200$ and $2200$) for evaluation.

\textbf{TANGO} dataset~\cite{tango} is introduced to support monocular vision-based spacecraft pose estimation. The dataset includes laboratory images of the Tango spacecraft model generated under diverse lighting conditions, backgrounds, and camera poses, with accurate bounding box, segmentation, and 6-DoF pose annotations. The dataset has been widely used to validate detection, segmentation, and pose estimation algorithms in spaceborne vision applications. The Tango dataset comprises $30,002$ training images and $3,002$ test images. We used the test split as is and randomly selected $500$ images from the training set.

\subsection{Metrics}
In line with previous related work~\cite{lin2014microsoft} on object detection and instance segmentation, we use Average Precision (AP) for evaluation.
AP is defined as the area under the precision–recall curve, averaged over IoU thresholds from $0.50$ to $0.95$ with a step size of $0.05$. In addition, AP$_{50}$ and AP$_{75}$ are commonly reported for reference. AP$_{50}$ uses a loose matching criterion (IoU $\geq$ 0.50), while AP$_{75}$ applies a stricter one (IoU $\geq$ 0.75). AP$_{50}$ mainly reflects detection capability, while AP$_{75}$ emphasizes localization accuracy. 

\subsection{Oracle}

To set the upper bound (Oracle) performance, we train two lightweight object detection models and one semantic segmentation model using ground-truth labels and evaluate them on the test sets. In our experiments, we adopt Efficient-Det~\cite{tan2020efficientdet} for object detection and YOLOv11~\cite{khanam2024yolov11} for object detection and segmentation, which have around $7M$ parameters and run in real time, i.e. both have FPS more than $60$. Both models are trained following the official implementations and the default training parameters for 300 epochs. During training and inference, the images are resized to $640\times640$ pixels for both models.

Table~\ref{tab:distill_res} presents the results of the Oracle experiment (see the first row of each dataset). 
Since the SPARK-2024 dataset has no segmentation labels, we train only the object detection models. 
Overall, both detectors achieve strong results across all datasets by achieving more than $95\%$ points on the challenging $AP_{75}$ metric, which is crucial for performing pose estimation or instance segmentation on cropped objects. YOLOv11 achieves the best overall performance on all datasets except for the SPEED+ Lightbox subset. 

\input{tables/zero_shot_res}

\subsection{Analysis of {\VFM}s on zero-shot detection}

We started our experiments by first evaluating the performance of {\VFM}s on three datasets. To this end, we selected three recent open-source {\VFM}s~\cite{zhang2023simple, zou2023segment, ren2024grounded} which achieve state-of-the-art results on open vocabulary detection and segmentation tasks on different benchmarks, and evaluated their zero-shot object detection performance on the curated test sets. Note that no fine-tuning is performed on these selected {\VFM}s using any form of labeled data.

Table~\ref{tab:zero_shot} presents the object detection and segmentation performances on the SPARK-2024, SPEED+, and TANGO datasets. We used \textit{spacecraft} as the textual prompt $T$ for each {\VFM} and gathered only the top prediction for each image. Prediction confidences are used directly to calculate the metrics for detection and segmentation. 

Except for the TANGO dataset, the SEEM and OpenSEED models achieve significantly low performance, especially on the detection task. 
OpenSEED achieves the best zero-shot segmentation result in the TANGO dataset, but its detection performance falls behind, especially on the $AP_{75}$ metric.
On the other hand, the GroundedSAM-2~\cite{ren2024grounded} model shows promising results, e.g., $90\%+$ on $AP_{50}$ detection task and $85\%+$ on $AP_{50}$ segmentation task on each dataset. 
Hence, we only used the GroundedSAM-2 model as the baseline {\VFM} in the next stages of our method. 

\input{tables/tta_res}

\subsection{Effect of TTA and WBF}

In order to improve the zero-shot detection performance of GroundedSAM-2, we experimented with different TTA settings. In natural images, the most common setting is to use the horizontal flip. Since space images have no notion of up and down, we added vertical flip to the augmentation list. Considering the nature of space images, we further introduced pixel manipulation augmentations such as brightness, saturation, and color to improve the overall contrast of the images. 

We ran extensive experiments on various combinations of augmentations on all datasets. In each case, the object detection predictions are merged using WBF with a threshold $\tau$ of $0.55$. The best results are achieved using only the vertical flip as an augmentation. Increasing the number of augmentations significantly decreased the detection and segmentation performance, most likely due to the possible low-confidence detections.

Table~\ref{tab:tta_wbf} presents the results when TTA and WBF are employed for zero-shot detection. In each dataset, we observed more than $1.5$ point improvement in the challenging $AP_{75}$ metric for the detection task. Similarly, an improvement of up to $1.0$ points is observed in the segmentation $AP_{50}$ metric.

\subsection{Distillation}

\input{tables/main_results}

In order to train the shallow detectors, we first extract the pseudo-labels (i.e. bounding boxes and segmentation masks) for the training sets following the same pipeline with TTA and WBF. Then, we use these pseudo-labels as ground-truth annotations and train the Efficient-Det and YOLOv11 models for both object detection and semantic segmentation tasks. We use the same hyper-parameters as \textit{Oracle} training, and similarly train the models for $300$ epochs.  
Table~\ref{tab:distill_res} summarizes the experimental results using TTA and WBF for each dataset (see the second rows as TTA+WBF). On the SPARK-2024 dataset, the YOLOv11 model significantly improves detection performance; notably, the metrics $AP$ and $AP_{75}$ are increased by more than $14$ and $25$, respectively. Similarly, TANGO and Lightbox detection performances have improved considerably on every metric. Regarding segmentation results, except for the Lightbox dataset, baseline zero-shot performance is greatly improved; e.g., on the TANGO dataset $AP$ and $AP_{75}$ metrics are increased by almost $10$ points.

Although the TTA improves the overall quality of the predictions, we observe that some low-confidence detections remain. Hence, before training the shallow models, we apply confidence filtering. 
We chose $0.5$ for the SPARK-2024 and TANGO datasets and $0.6$ for the SPEED+ dataset as confidence thresholds to remove low-quality detections. 
These values are selected on the basis of the mean confidences of the predictions for each dataset.  
After this step, the number of training images dropped to 395, 451, 417, and 471 for SPARK-2024, Sunlamp, Lightbox, and TANGO, respectively. We retrain detection and segmentation models in these confidence-filtered training sets. 
The results are presented in Table~\ref{tab:distill_res} in the third rows (marked as TTA+WBF+CF) of each dataset. We observe a consistent improvement on all datasets for both tasks. Notably, the detection performance of the SPARK-2024 dataset is increased by $8$ points on the challenging $AP_{75}$ metric. Similarly, TANGO segmentation metrics are improved up to $3$ points.  

\input{figures/visual_results}
Here we take a step forward and employ the iterative knowledge distillation, i.e. using the current student model as the next teacher. 
We relabel the training sets using the trained student model and train the student model from scratch. 
Table~\ref{tab:distill_res} presents the results of this iterative distillation experiment in the last rows (denoted TTA + WBF + CF + TR).
In the SPARK-2024 dataset, the detection $AP$ is improved by more than $5$ points. On other datasets, we observe up to $3$ point improvements for different metrics both for detection and segmentation tasks. However, in rare cases, there happens a small drop in performance, e.g., segmentation results of the Sunlamp dataset, most likely due to overfitting.

Compared to the initial {\VFM} predictions, our proposed pipeline significantly improves (e.g., more than $10$ points on TANGO segmentation $AP$) the detection and segmentation results on three challenging datasets, even with very shallow student models.

\subsection{Visual Results}

In Figure~\ref{fig:visuals}, we present visual results containing the input image, ground-truth labels, zero-shot {\VFM} predictions, and our student model predictions for each dataset. Our method improves both detection and segmentation performance even when the pseudo-labels contain significant noise, as in the cases of the TANGO dataset. These results show the importance of label distillation in improving performance.

%% file: tables/zero_shot_res.tex
\begin{table}
\caption{Zero-shot Detection and Segmentation Performances of {\VFM}s on the SPARK-2024, SPEED+ and TANGO Test Sets.}
\vspace{-7mm}
\begin{center}
\resizebox{1.0\columnwidth}{!}{
\begin{tabular}{llccc|ccc}
\toprule 
 & \multirow{3}{*}{VFM}  & \multicolumn{3}{c}{Detection} & \multicolumn{3}{c}{Segmentation} \\
\cmidrule(lr){3-5} \cmidrule(lr){6-8}
 &  & $AP$ & $AP_{50}$ &  $AP_{75}$  & $AP$ & $AP_{50}$ &  $AP_{75}$ \\
\midrule 
\parbox[t]{2mm}{\multirow{3}{*}{\rotatebox[origin=c]{90}{Spark}}} 
& SEEM  & 17.7 & 49.8 & 2.6             & N/A & N/A &	N/A  \\
& OpenSEED  & 8.5 & 28.0 & 0.8          & N/A & N/A &	N/A \\
& GSAM-2  & 53.3	& 97.9	& 61.7      & N/A & N/A &	N/A  \\
\midrule 
\parbox[t]{2mm}{\multirow{3}{*}{\rotatebox[origin=c]{90}{SLamp}}} 
& SEEM  & 25.3	& 45.4	& 23.7            & 28.0 & 43.6 &	31.1 \\
& OpenSEED  & 33.1	& 63.3	& 32.1        & 40.2 & 62.8 &	45.3  \\
& GSAM-2  & 73.9	& 97.9	& 92.1        & 72.5 & 94.7 &	85.0 \\
\midrule 
\parbox[t]{2mm}{\multirow{3}{*}{\rotatebox[origin=c]{90}{LBox}}} 
& SEEM      & 11.4	& 19.6  & 11.3      & 37.1 & 52.6 &	41.8 \\
& OpenSEED  & 35.0	& 59.3	& 36.9      & 37.1 & 54.8 &	41.6 \\
& GSAM-2    & 65.8	& 91.4	& 74.0      & 66.2 & 83.9 &	73.5 \\
\midrule 
\parbox[t]{2mm}{\multirow{3}{*}{\rotatebox[origin=c]{90}{Tango}}} 
& SEEM      & 25.7 & 60.5 &	7.7  & 41.4 & 69.7 & 41.5 \\
& OpenSEED  & 23.8 & 87.7 &	0.4 & 62.2 & 91.6 &	77.9 \\
& GSAM-2    & 58.2 & 91.8 &	75.7  & 58.2 & 85.9 & 69.0 \\
\bottomrule 
\end{tabular}}
\end{center}
\label{tab:zero_shot}
\vspace{-5mm}
\end{table}

%% file: tables/tta_res.tex
\begin{table}
\caption{The Effect of Utilizing TTA and WBF with GroundedSAM-2.}
\vspace{-4mm}
\begin{center}
\resizebox{1.0\columnwidth}{!}{
\begin{tabular}{lccc|ccc}
\toprule 
 \multirow{3}{*}{Dataset}  & \multicolumn{3}{c}{Detection} & \multicolumn{3}{c}{Segmentation} \\
\cmidrule(lr){2-4} \cmidrule(lr){5-7}
 & $AP$ & $AP_{50}$ &  $AP_{75}$  & $AP$ & $AP_{50}$ &  $AP_{75}$ \\
\midrule 


Spark   & 54.4 & 98.0 & 64.0 & N/A & N/A & N/A \\
SLamp   & 74.6 & 98.9 & 93.7 & 73.2 & 95.8 & 86.3 \\
LBox    & 66.9 & 91.7 & 75.6 & 66.8 & 85.0 & 73.8 \\
Tango   & 58.5 & 92.0 & 76.3 & 58.7 & 86.4 & 69.6 \\

\bottomrule 
\end{tabular}}
\end{center}
\label{tab:tta_wbf}
\vspace{-6mm}
\end{table}

%% file: tables/main_results.tex
\begin{table*}[t]
\caption{Oracle and Distillation Results on the SPARK, SPEED+ and TANGO Test Sets. CF: Confidence Filtering, TR: Train Relabeling.}
\vspace{-4mm}
\label{tab:distill_res}
\begin{center}
\resizebox{0.90\textwidth}{!}{
\begin{tabular}{llcccccc|ccc}
\toprule 
 & \multirow{5}{*}{Method}  & \multicolumn{6}{c}{Detection} & \multicolumn{3}{c}{Segmentation} \\
\cmidrule(lr){3-8} \cmidrule(lr){9-11}
&  & \multicolumn{3}{c}{Efficient-Det} & \multicolumn{3}{c}{YOLOv11} & \multicolumn{3}{c}{YOLOv11} \\
 \cmidrule(lr){3-5} \cmidrule(lr){6-8} \cmidrule(lr){9-11}
 &  & $AP$ & $AP_{50}$ &  $AP_{75}$  & $AP$ & $AP_{50}$ &  $AP_{75}$ & $AP$ & $AP_{50}$ &  $AP_{75}$ \\
\midrule 
\parbox[t]{2mm}{\multirow{4}{*}{\rotatebox[origin=c]{90}{Spark}}} 
& Oracle        & 98.7  & 99.0 & 99.0	  & 99.2 & 99.5	& 99.5          & N/A & N/A &	N/A  \\
& TTA+WBF       & 65.0  & 98.9 & 78.9	  & 68.9 & 99.5 & 89.4          & N/A & N/A &	N/A \\
& TTA+WBF+CF    & 66.8  & 98.9 & 80.4     & 70.8 & 99.5 & 97.4          & N/A  & N/A  & N/A  \\
& TTA+WBF+CF+TR & 68.7  & 98.9 & 80.5     & 76.4 & 99.5 & 98.2          & N/A  & N/A  & N/A  \\
\midrule 
\parbox[t]{2mm}{\multirow{4}{*}{\rotatebox[origin=c]{90}{Sunlamp}}} 
& Oracle        & 81.6	& 99.0	& 97.8      & 85.0	& 99.5	& 98.8      & 81.5 & 94.3 &	94.1  \\
& TTA+WBF       & 75.4  & 99.0	& 94.9      & 74.2	& 95.4	& 92.0      & 73.0 & 94.4 &	91.0 \\
& TTA+WBF+CF    & 75.0	& 99.0	& 94.7      & 74.6	& 95.3	& 91.9      & 73.0 & 94.8 &	91.6  \\
& TTA+WBF+CF+TR & 74.9	& 99.0	& 94.7      & 75.0	& 95.3	& 92.6      & 74.0 & 94.3 &	91.0  \\
\midrule 
\parbox[t]{2mm}{\multirow{4}{*}{\rotatebox[origin=c]{90}{Lightbox}}} 
& Oracle        & 84.7 & 99.0	& 97.6      & 80.3	& 99.4	& 93.4      & 82.3 & 99.5 &	95.7   \\
& TTA+WBF       & 66.3	& 94.9	& 76.0      & 69.1	& 96.4	& 79.3      & 61.6 & 91.1 & 72.6  \\
& TTA+WBF+CF    & 66.4	& 97.0	& 76.8      & 69.6	& 96.5	& 80.1      & 62.5 & 92.1 &	72.6   \\
& TTA+WBF+CF+TR & 68.2	& 97.2	& 79.1      & 71.0  & 96.8	& 81.6      & 63.5 & 93.6 &	73.7   \\
\midrule 
\parbox[t]{2mm}{\multirow{4}{*}{\rotatebox[origin=c]{90}{Tango}}} 
& Oracle        & 87.4 & 94.8   & 93.0      & 92.1  & 99.4  & 97.7      & 79.1 & 99.1 &	96.2  \\
& TTA+WBF       & 66.0 & 92.2   & 87.0      & 85.0  & 93.5  & 93.1      & 67.0 & 93.4 &	84.1 \\
& TTA+WBF+CF    & 66.2 & 92.6	& 89.0      & 86.4  & 94.9  & 93.7      & 69.7 & 94.5 &	87.1  \\
& TTA+WBF+CF+TR & 70.6 & 92.1	& 90.4      & 86.5  & 95.6  & 93.8               & 70.0 & 94.7 &	86.9  \\
\bottomrule 
\end{tabular}}
\end{center}
\vspace{-5mm}
\end{table*}

%% file: figures/visual_results.tex
\begin{figure*}[t]
\centering
\setlength\tabcolsep{1.0pt} 
\begin{tabular}{lcccc}

{\textit{\textbf{{\small}}}} & {\textit{\textbf{{Input}}}} & {\textit{\textbf{{GT}}}} & {\textit{\textbf{{{\VFM}}}}} & {\textit{\textbf{{Ours}}}}  \\ 

{\rotatebox[origin=t]{90}{\textit{\textbf{SPARK}}}}  &
\includegraphics[width=\myw,  ,valign=m, keepaspectratio,] {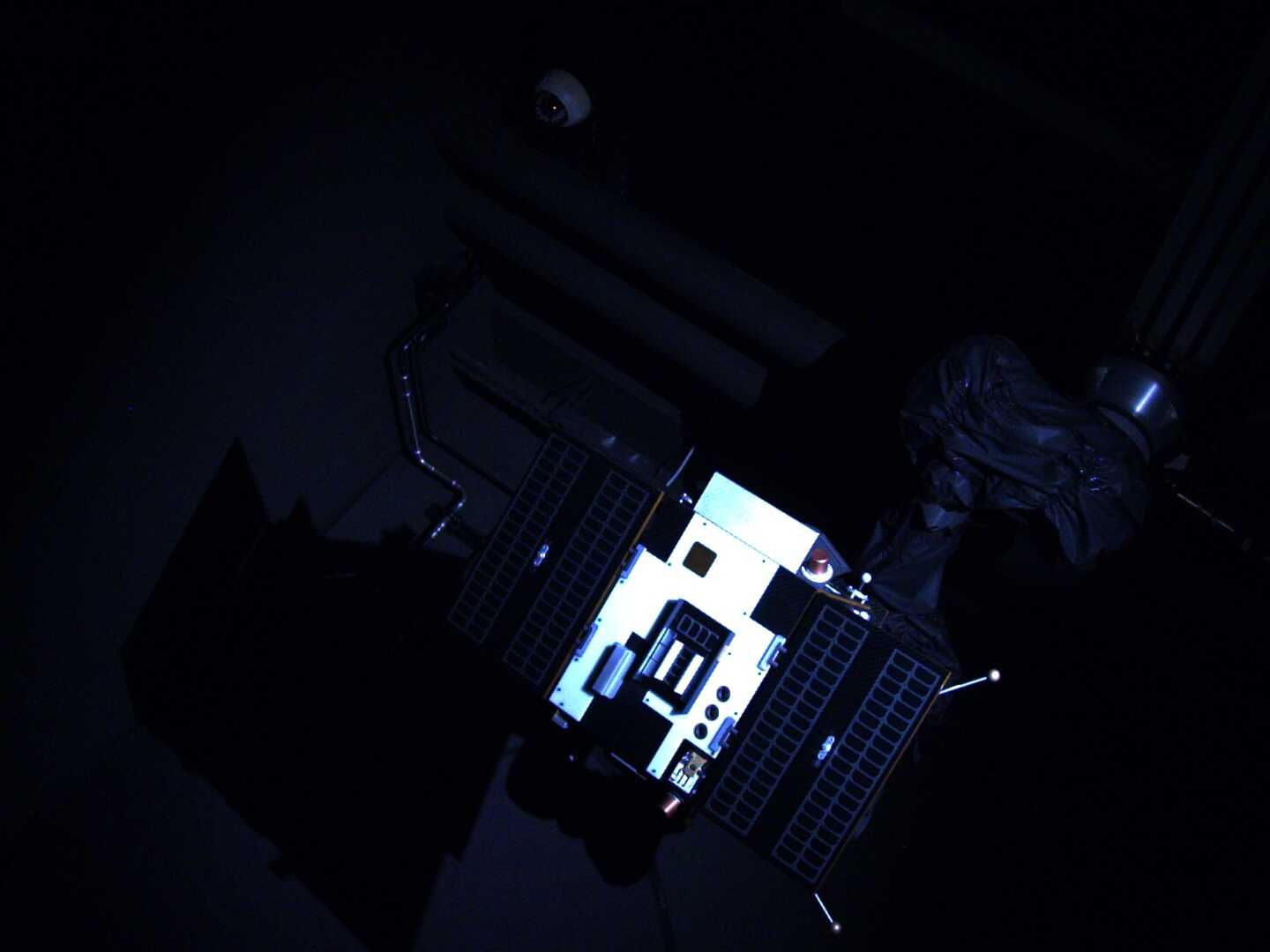} &
\includegraphics[width=\myw,  ,valign=m, keepaspectratio,] {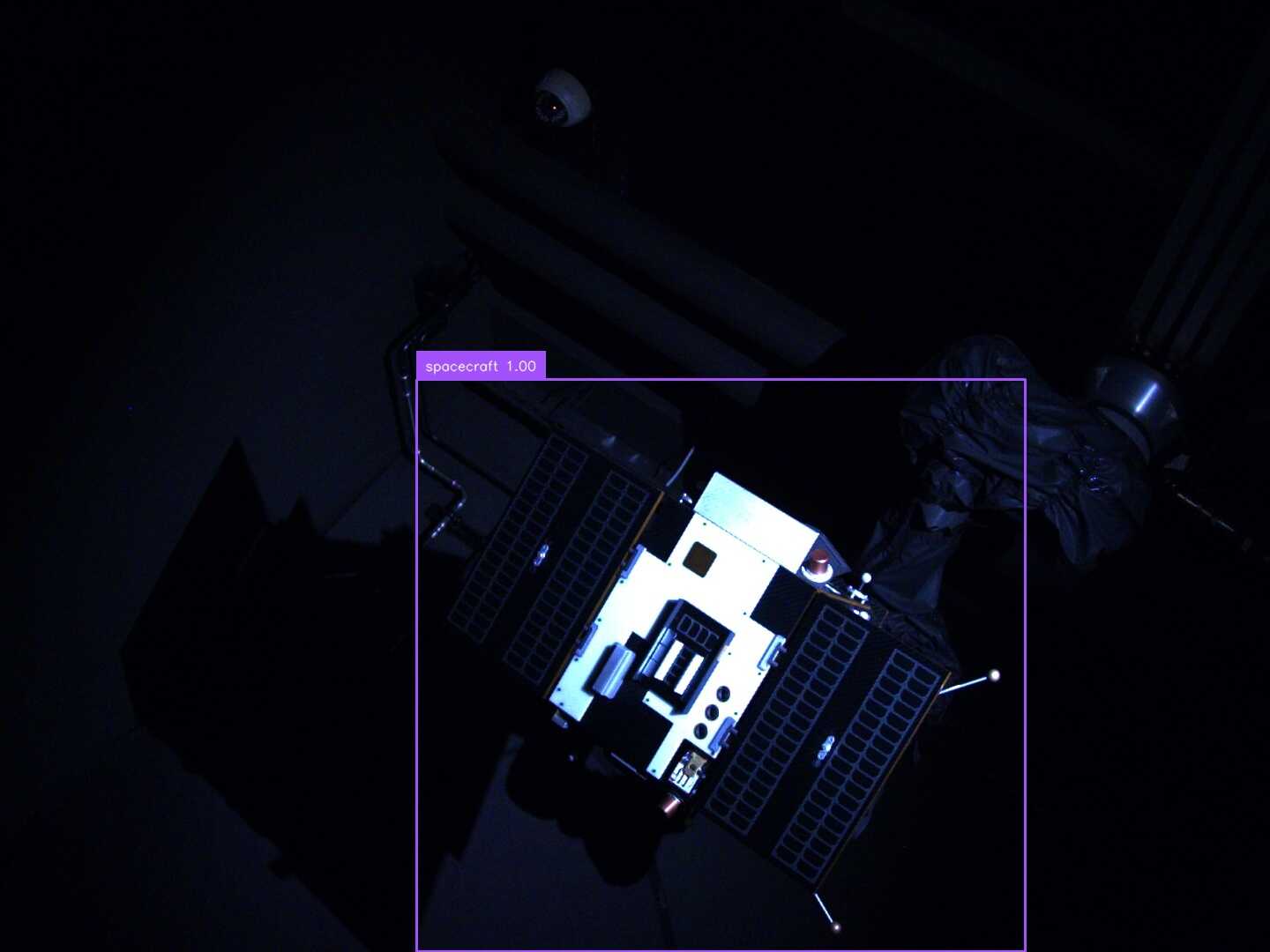} &
\includegraphics[width=\myw,  ,valign=m, keepaspectratio,] {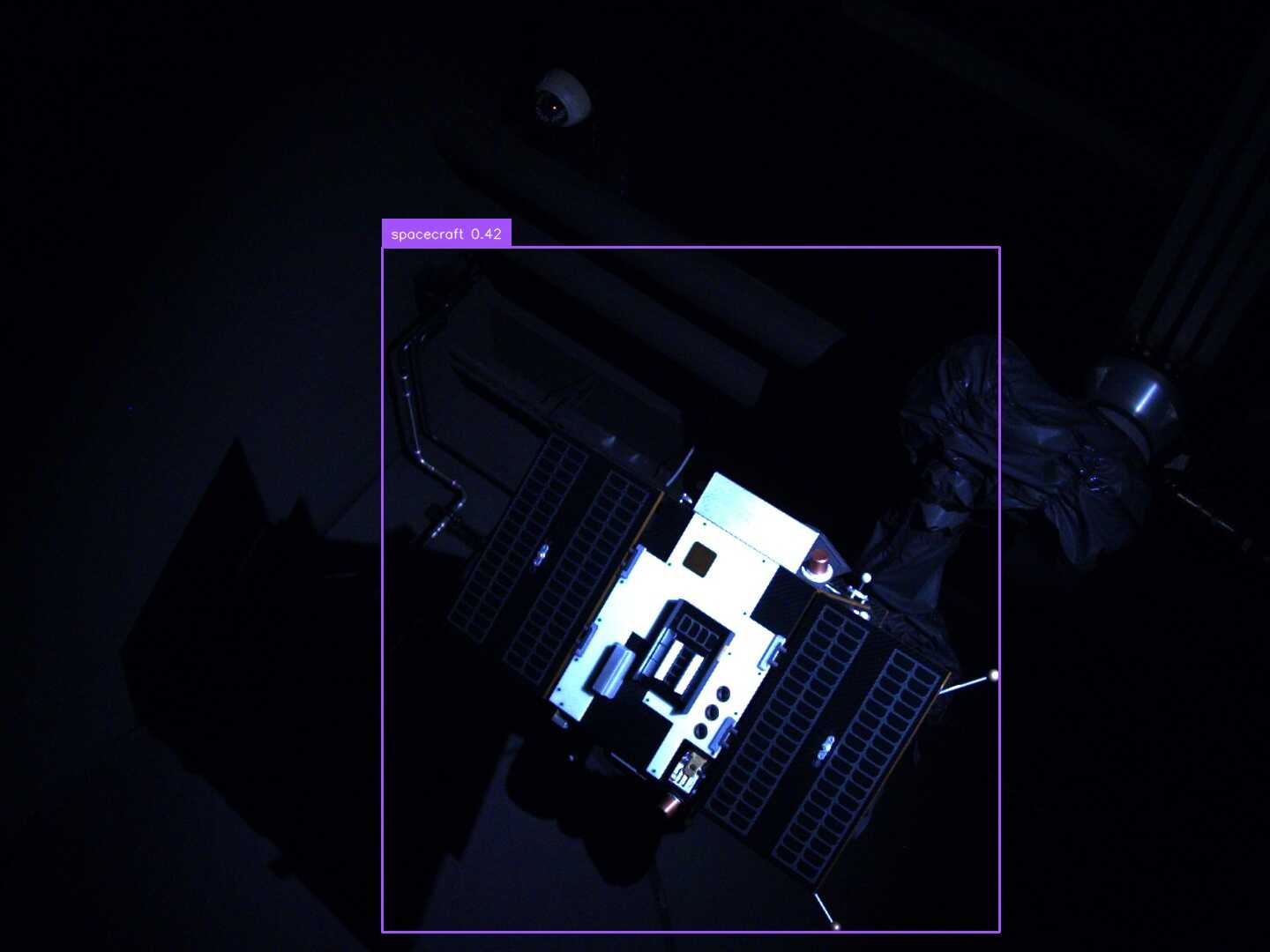} & 
\includegraphics[width=\myw,  ,valign=m, keepaspectratio,] {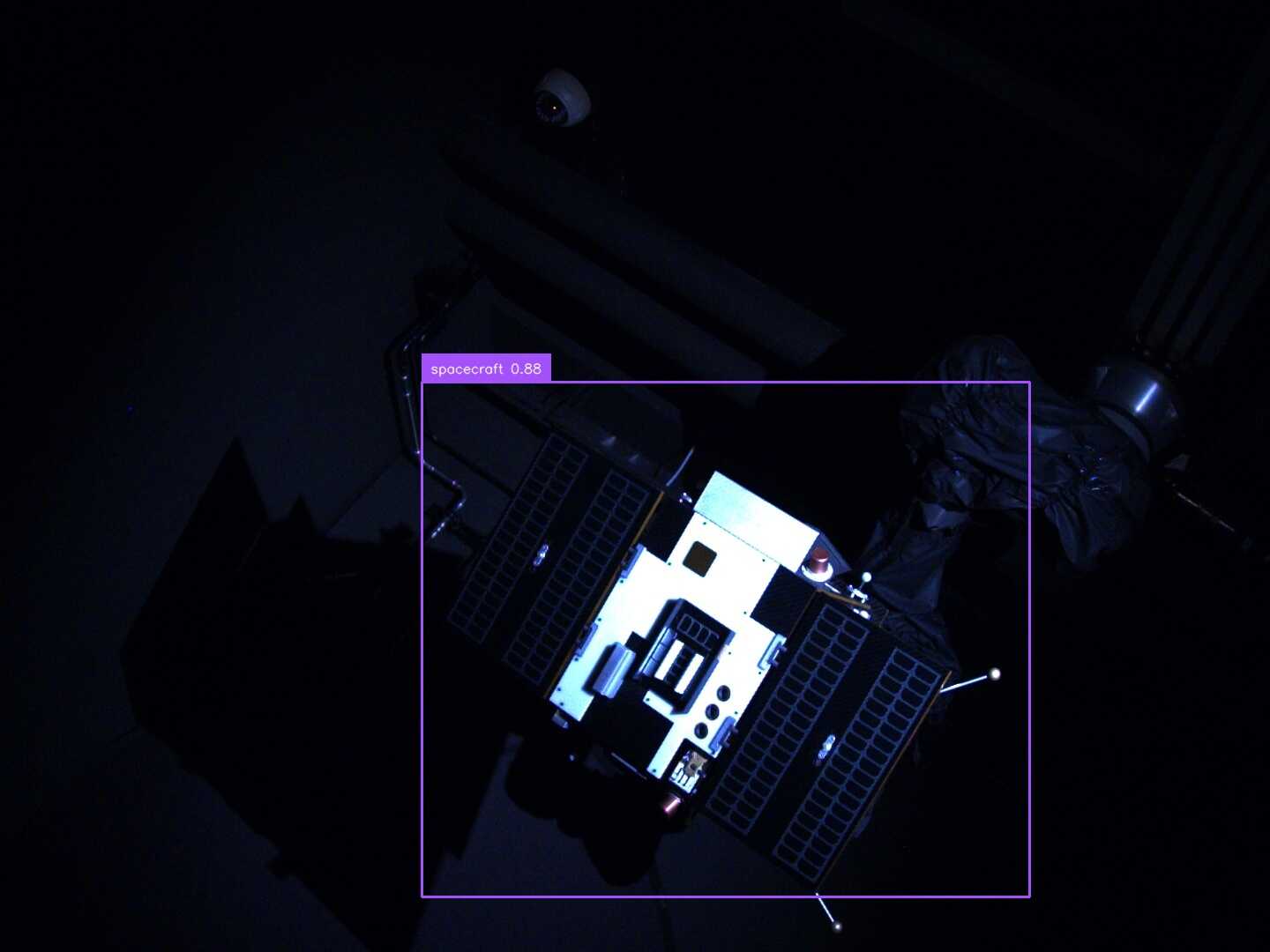}\\

{\rotatebox[origin=t]{90}{\textit{\textbf{SPARK}}}}  &
\includegraphics[width=\myw,  ,valign=m, keepaspectratio,] {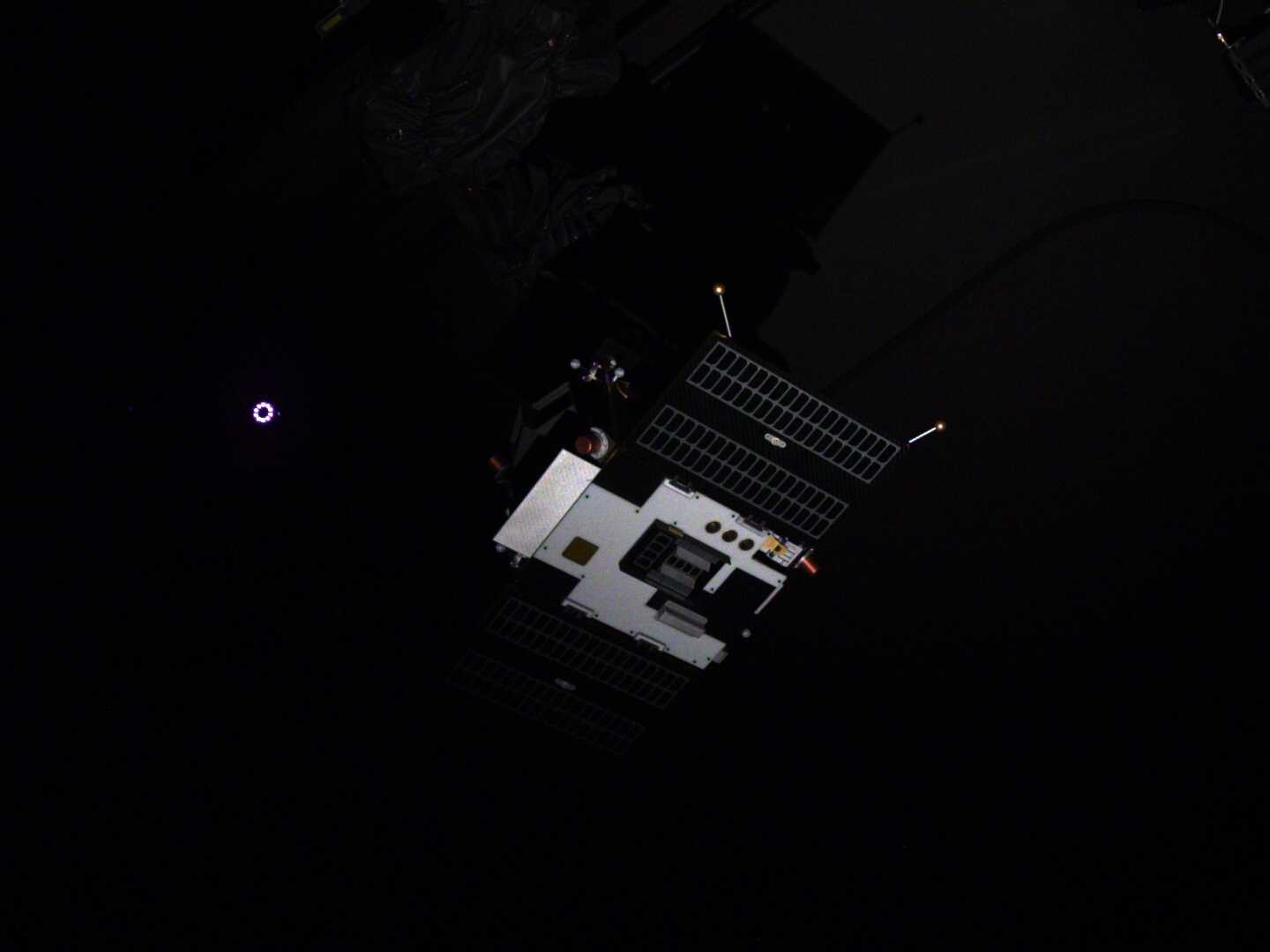} &
\includegraphics[width=\myw,  ,valign=m, keepaspectratio,] {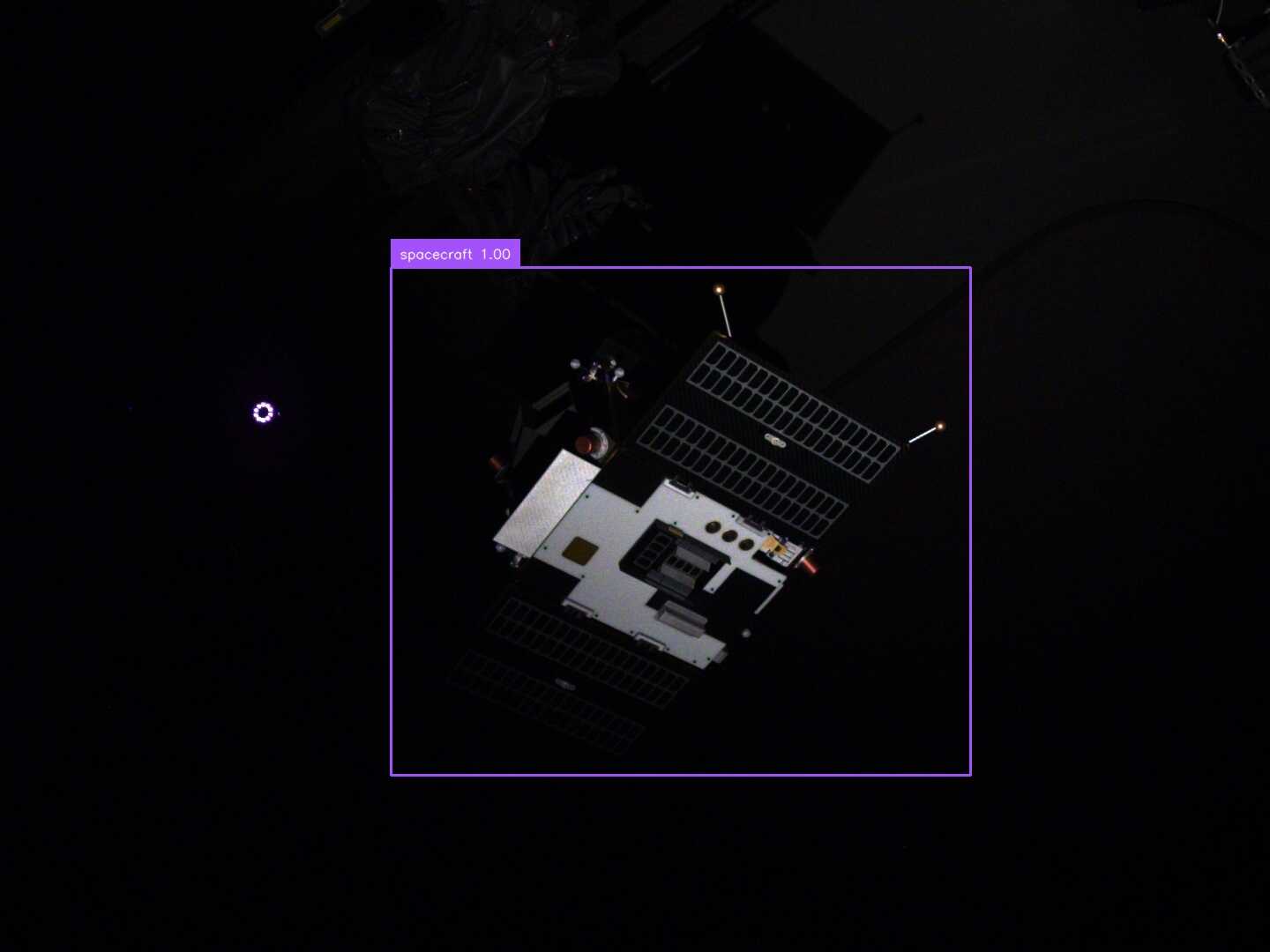} &
\includegraphics[width=\myw,  ,valign=m, keepaspectratio,] {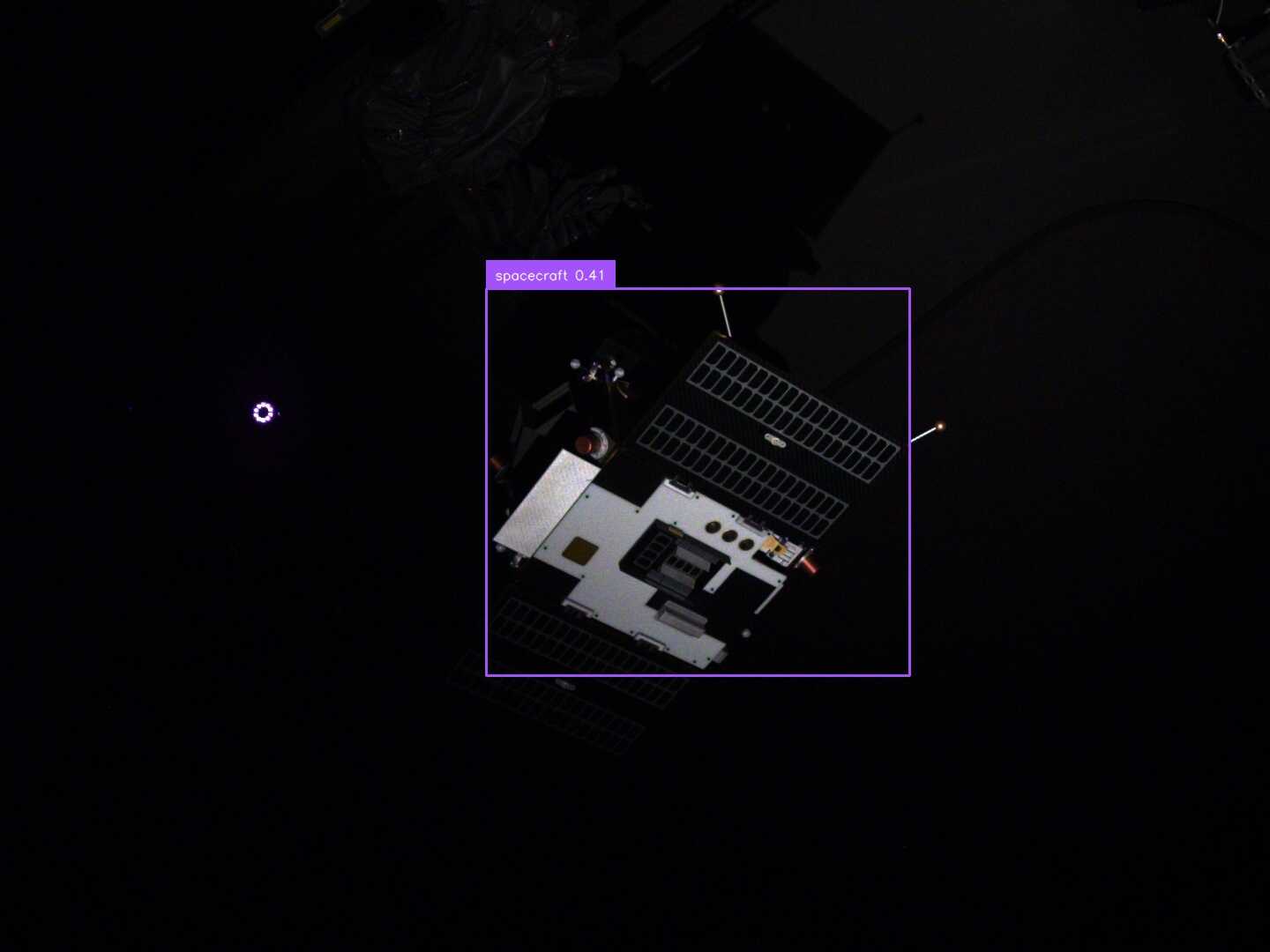} & 
\includegraphics[width=\myw,  ,valign=m, keepaspectratio,] {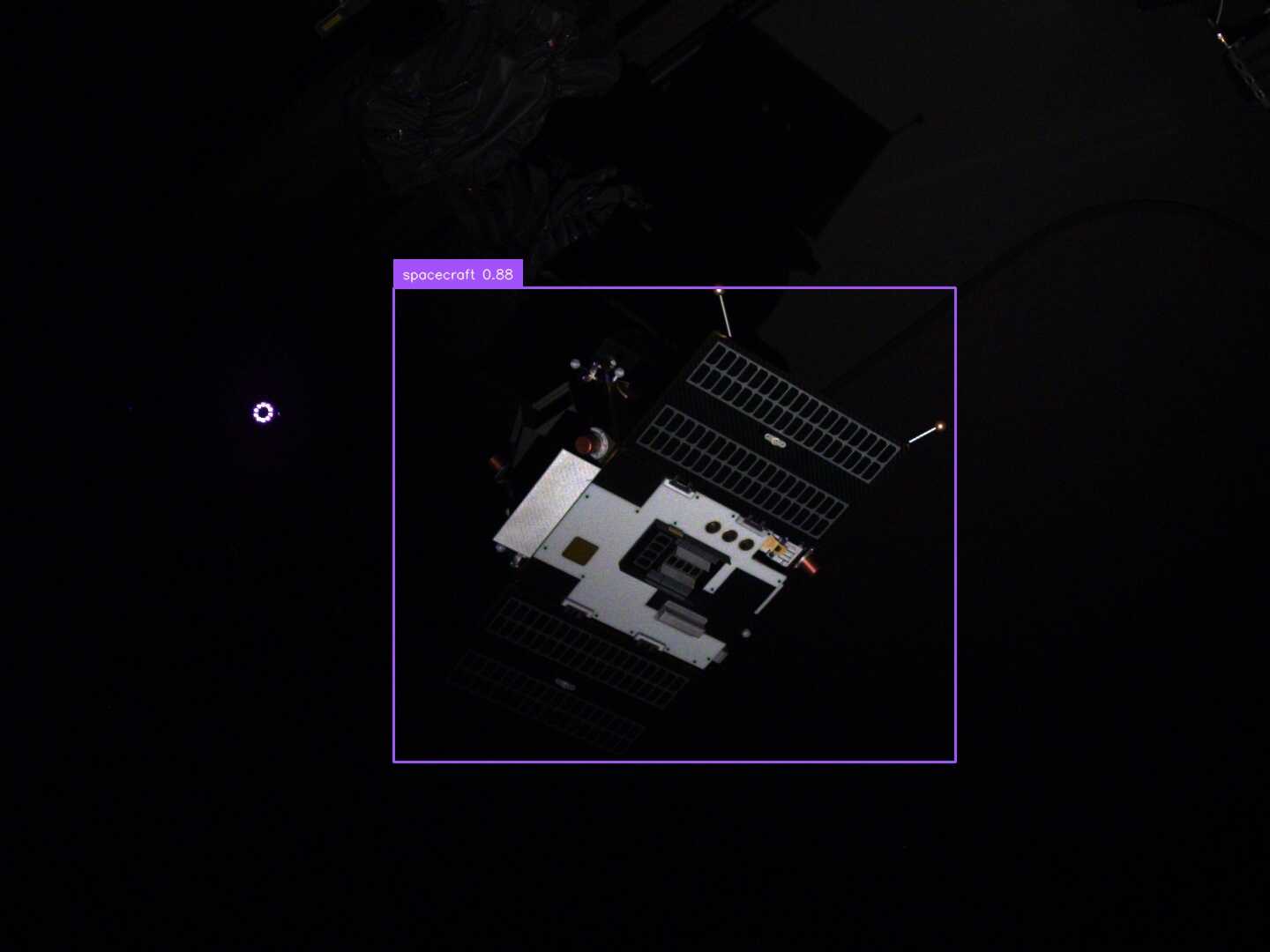}\\

{\rotatebox[origin=t]{90}{\textit{\textbf{LightBox}}}}  &
\includegraphics[width=\myw,  ,valign=m, keepaspectratio,] {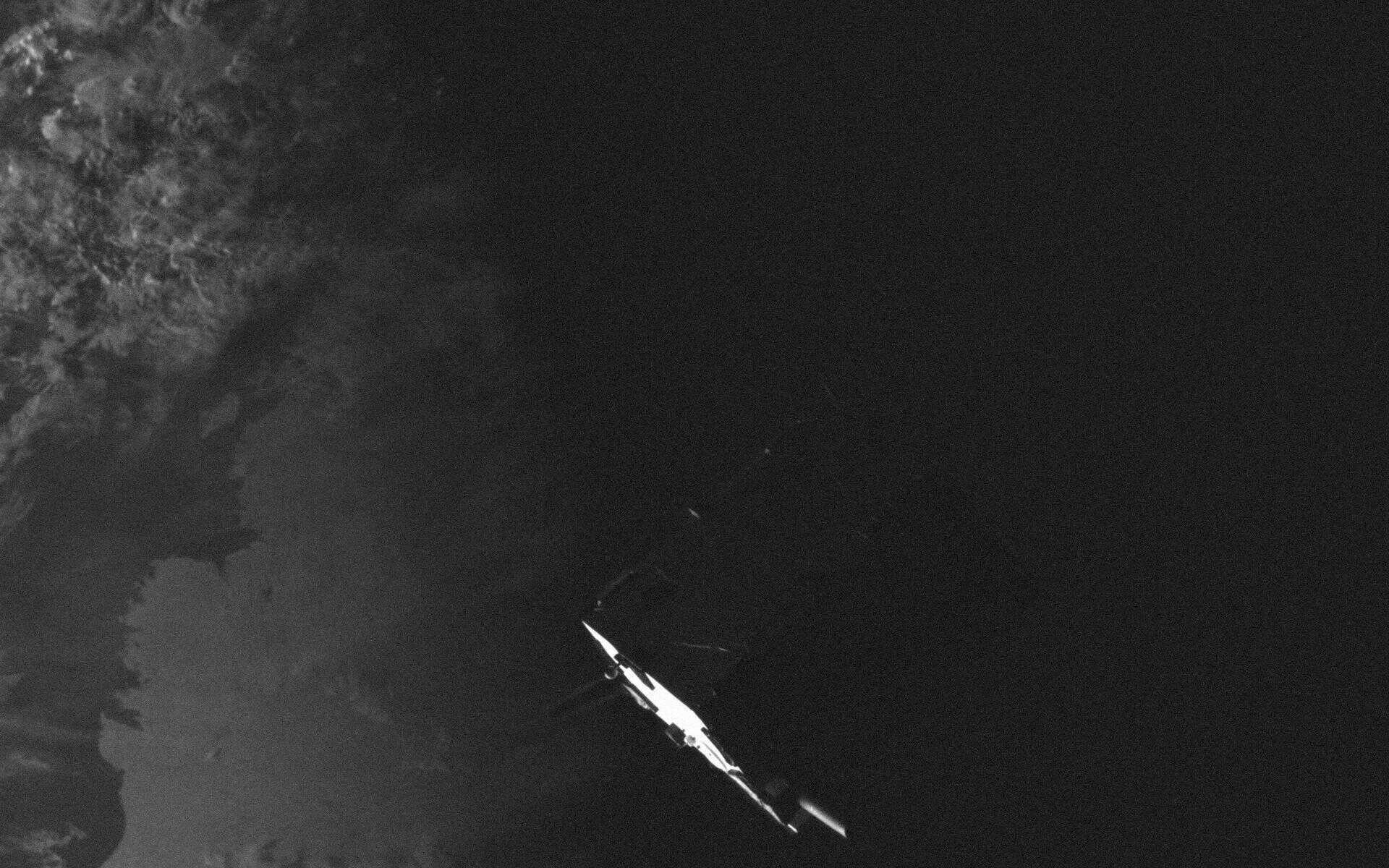} &
\includegraphics[width=\myw,  ,valign=m, keepaspectratio,] {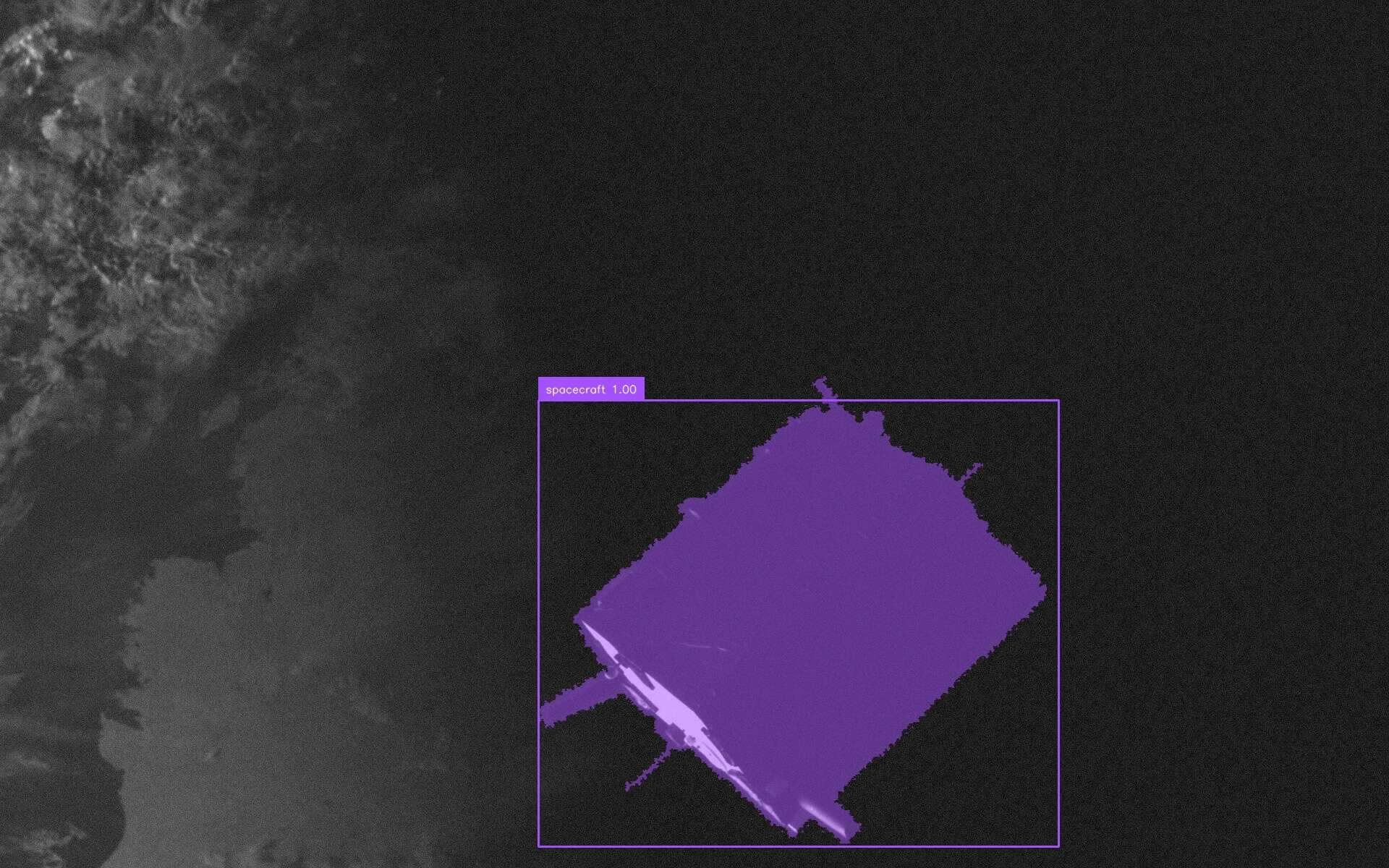} &
\includegraphics[width=\myw,  ,valign=m, keepaspectratio,] {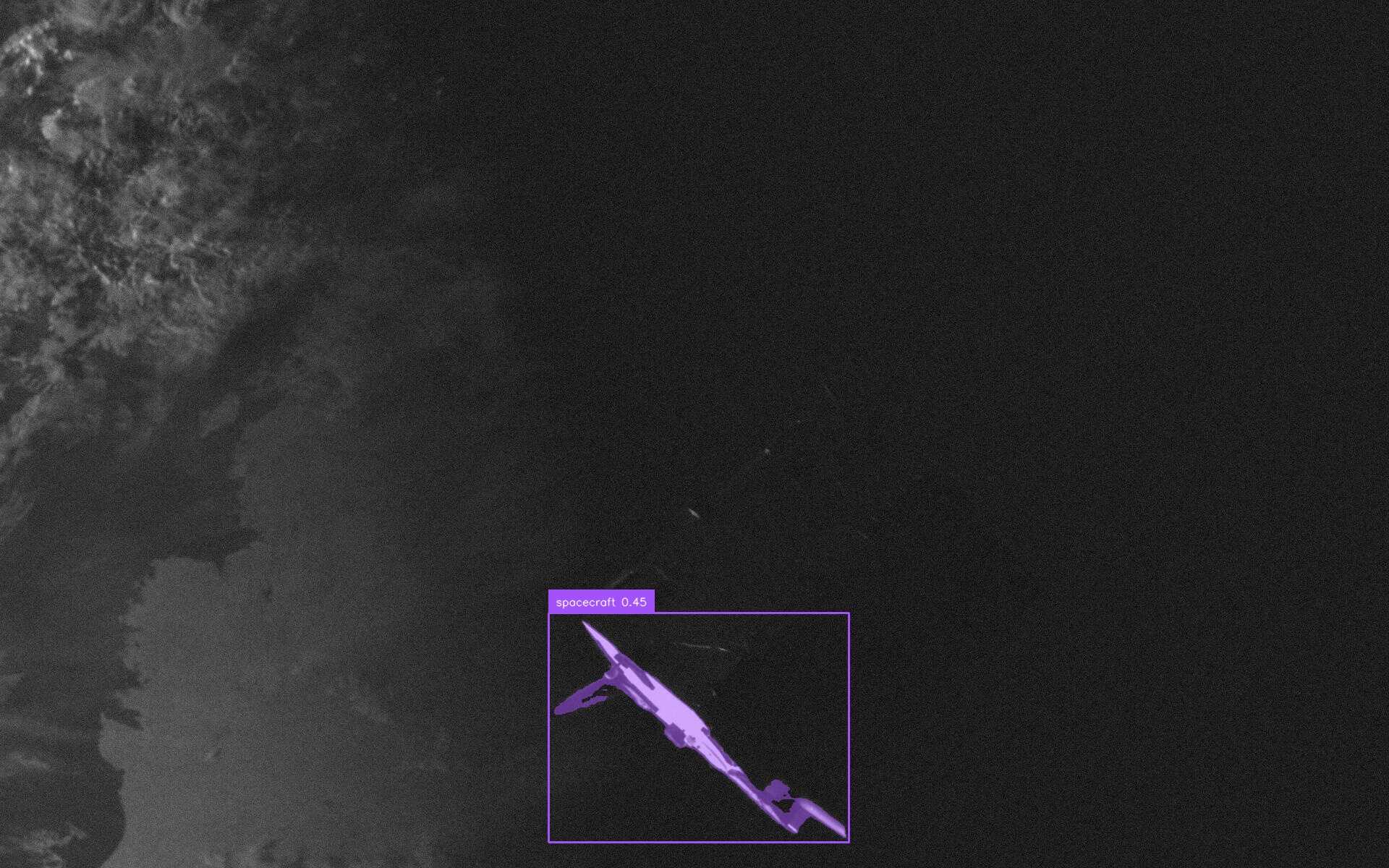} & 
\includegraphics[width=\myw,  ,valign=m, keepaspectratio,] {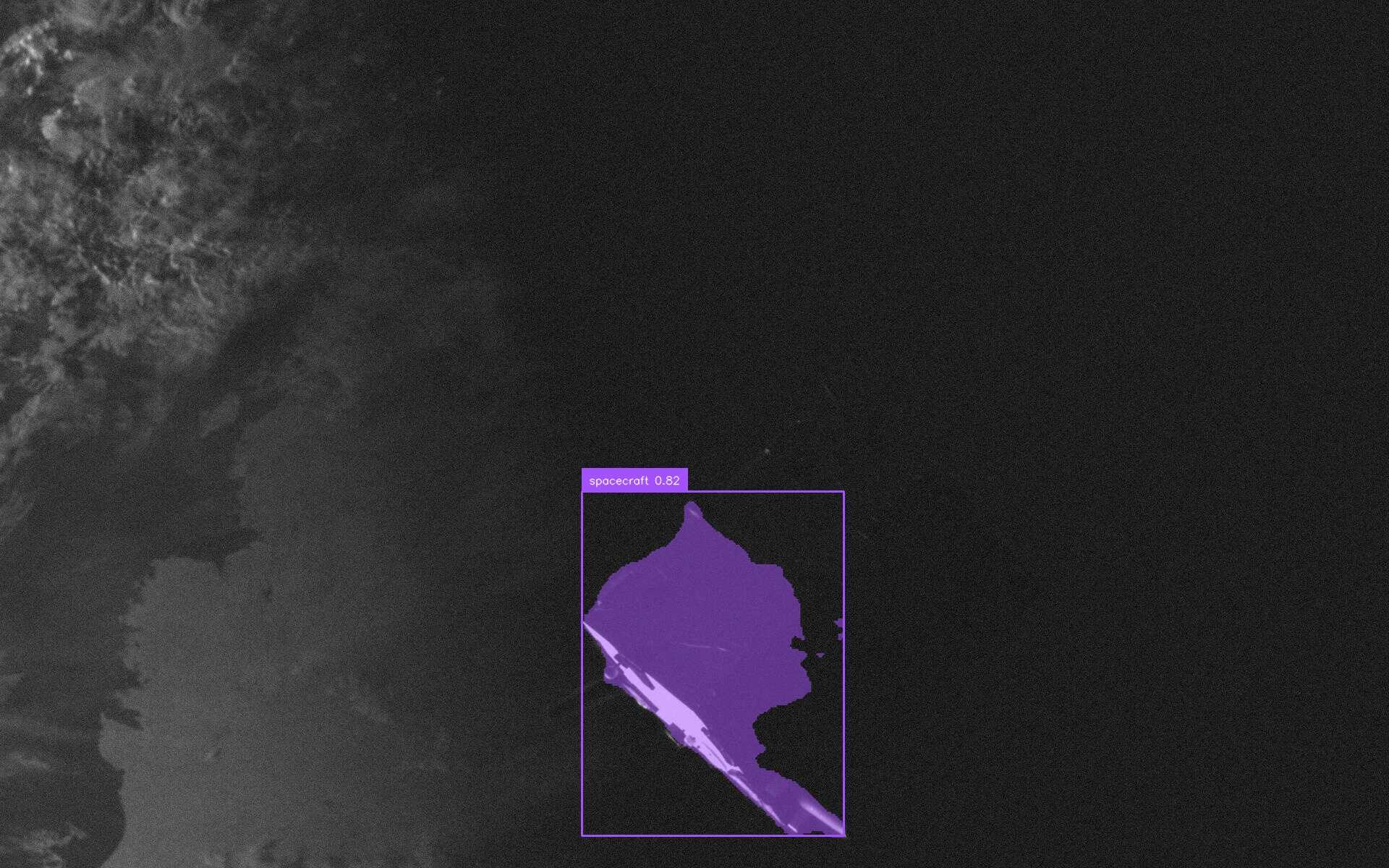}\\

{\rotatebox[origin=t]{90}{\textit{\textbf{LightBox}}}}  &
\includegraphics[width=\myw,  ,valign=m, keepaspectratio,] {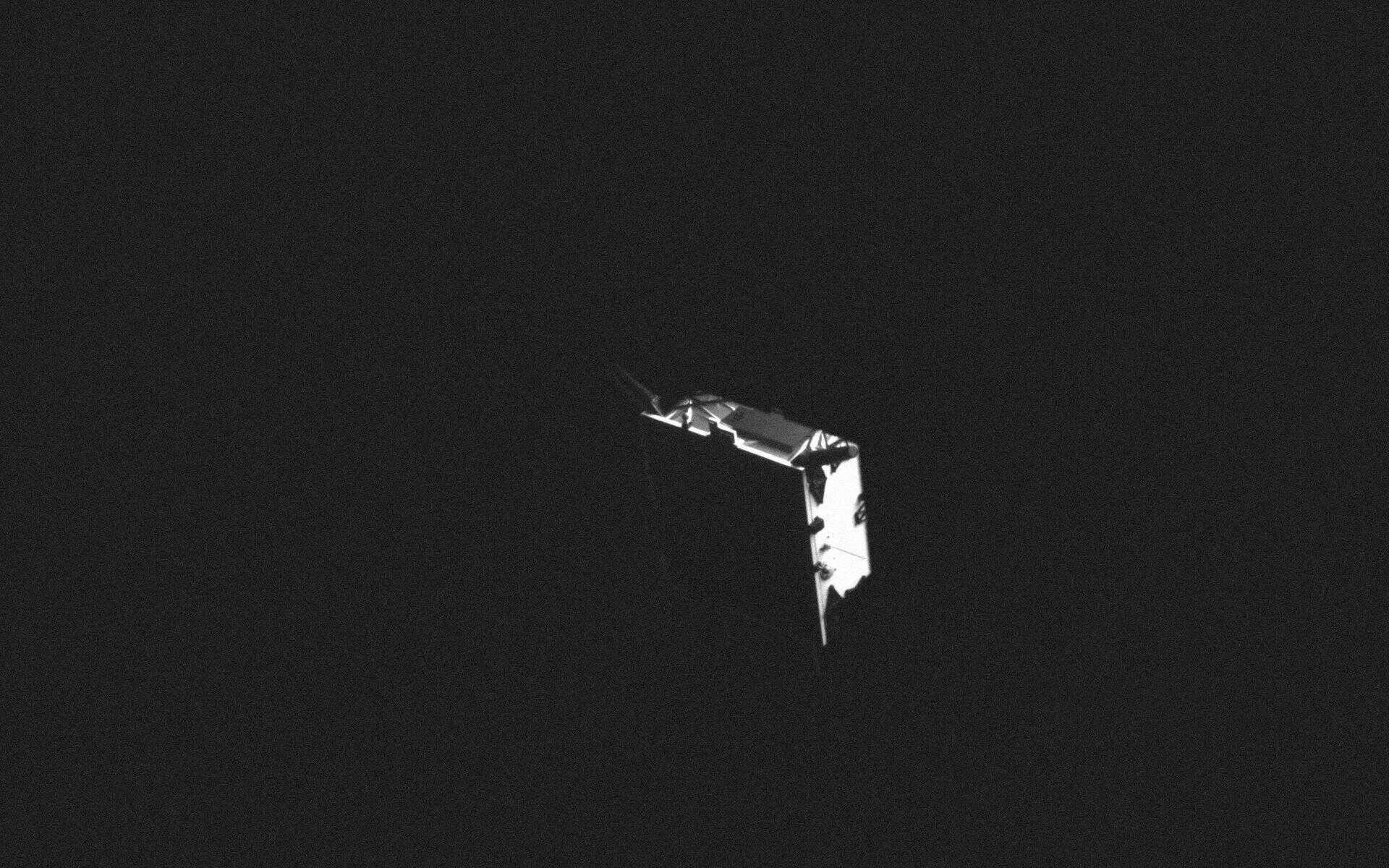} &
\includegraphics[width=\myw,  ,valign=m, keepaspectratio,] {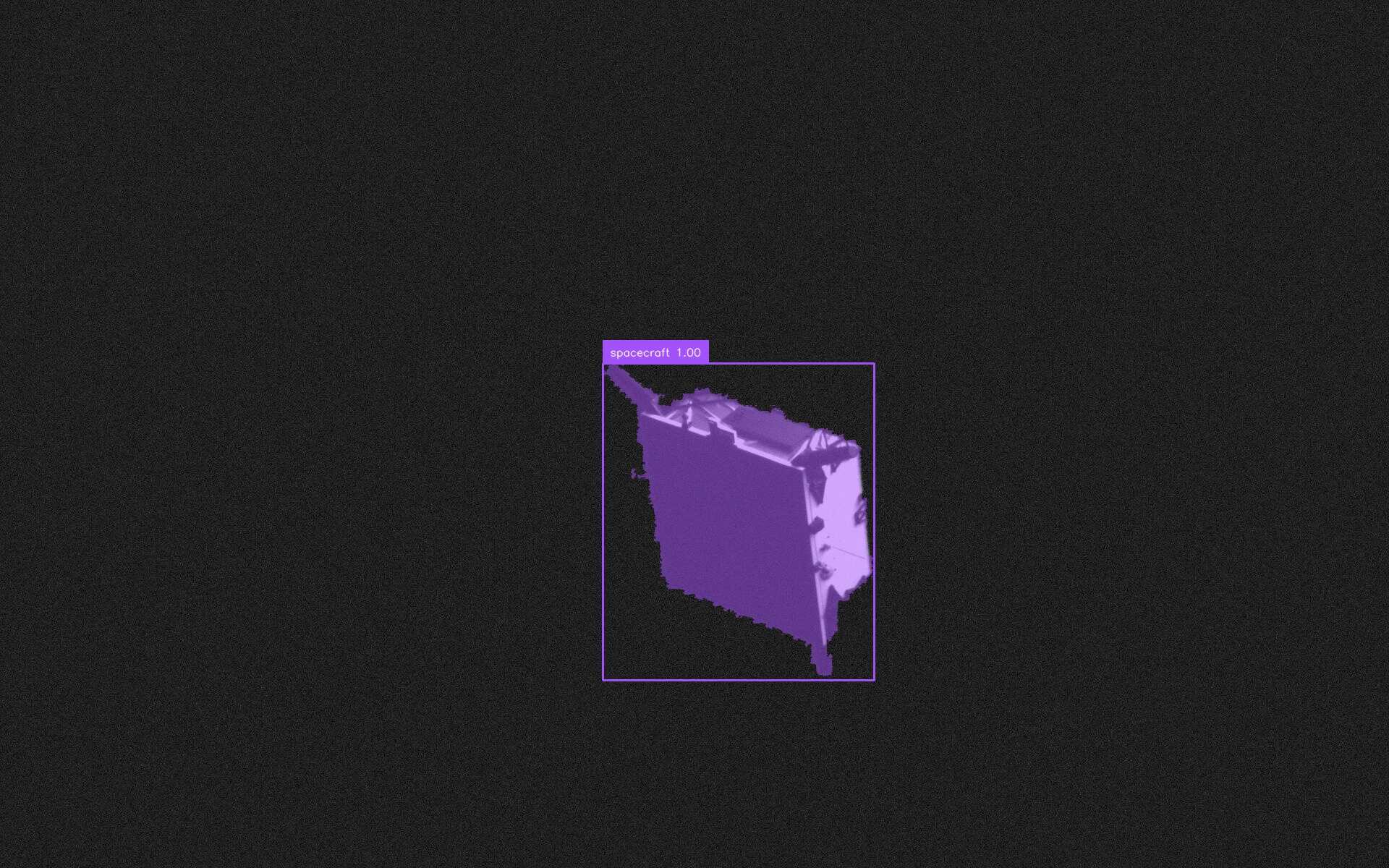} &
\includegraphics[width=\myw,  ,valign=m, keepaspectratio,] {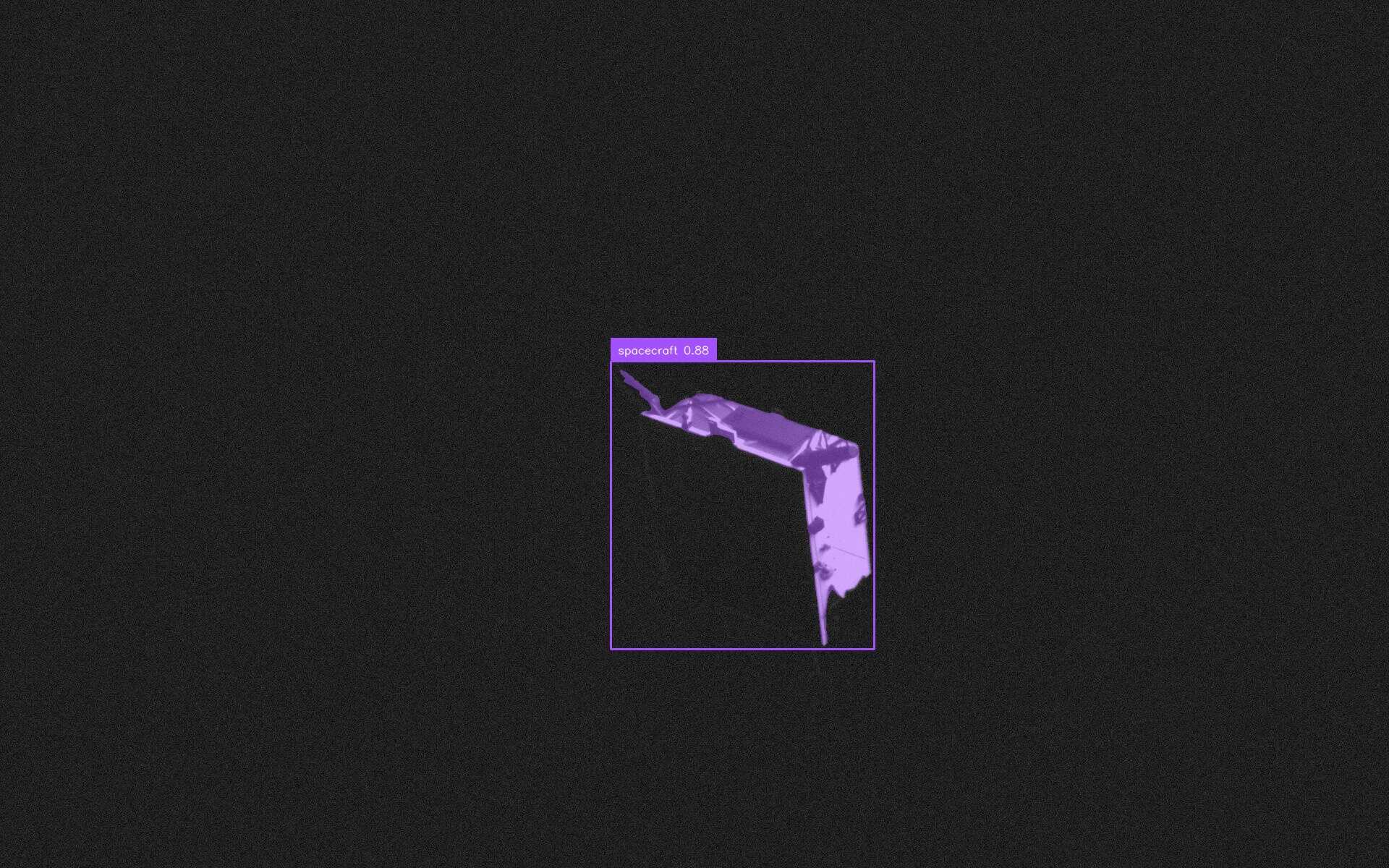} & 
\includegraphics[width=\myw,  ,valign=m, keepaspectratio,] {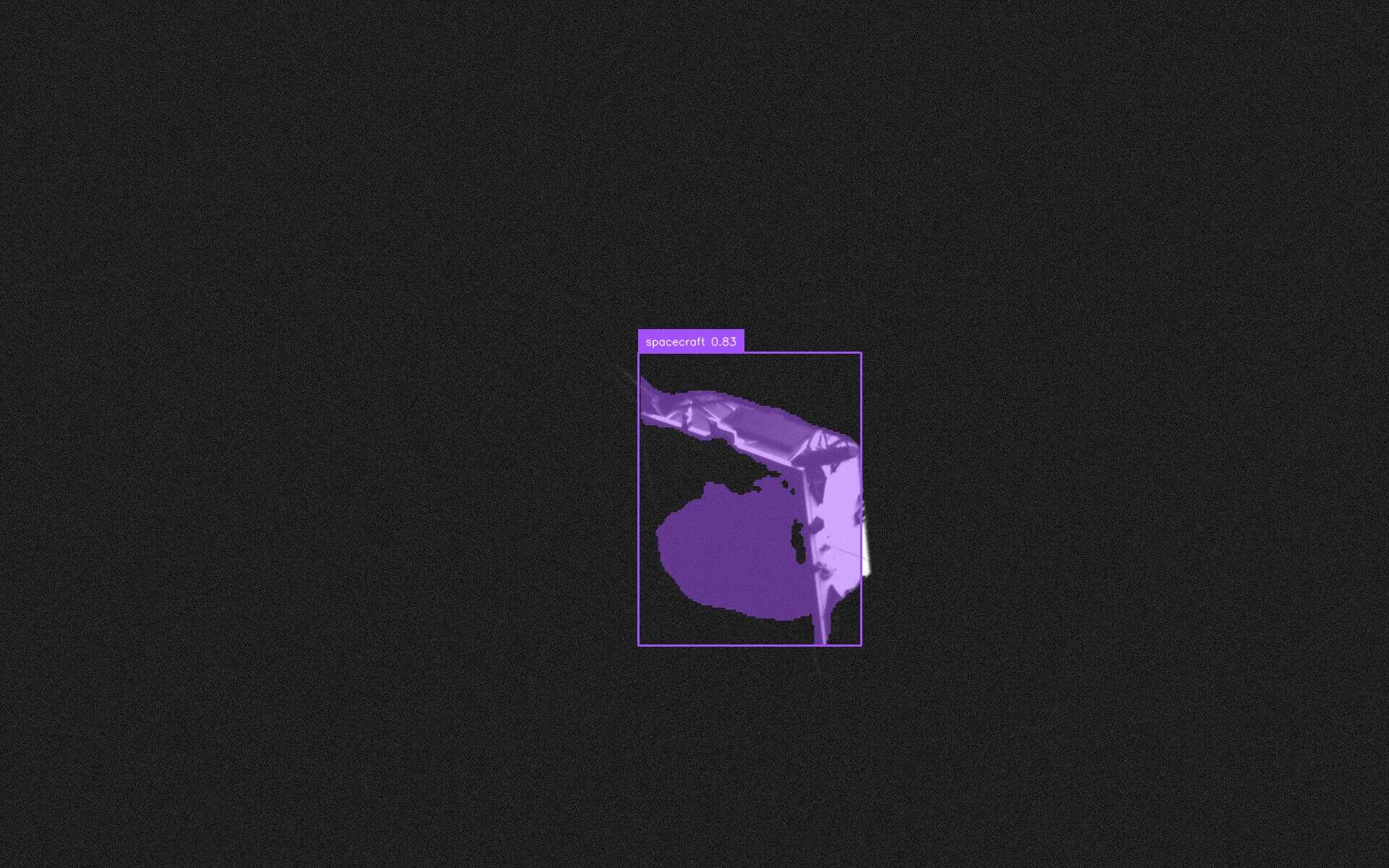}\\

{\rotatebox[origin=t]{90}{\textit{\textbf{SunLamp}}}}  &
\includegraphics[width=\myw,  ,valign=m, keepaspectratio,] {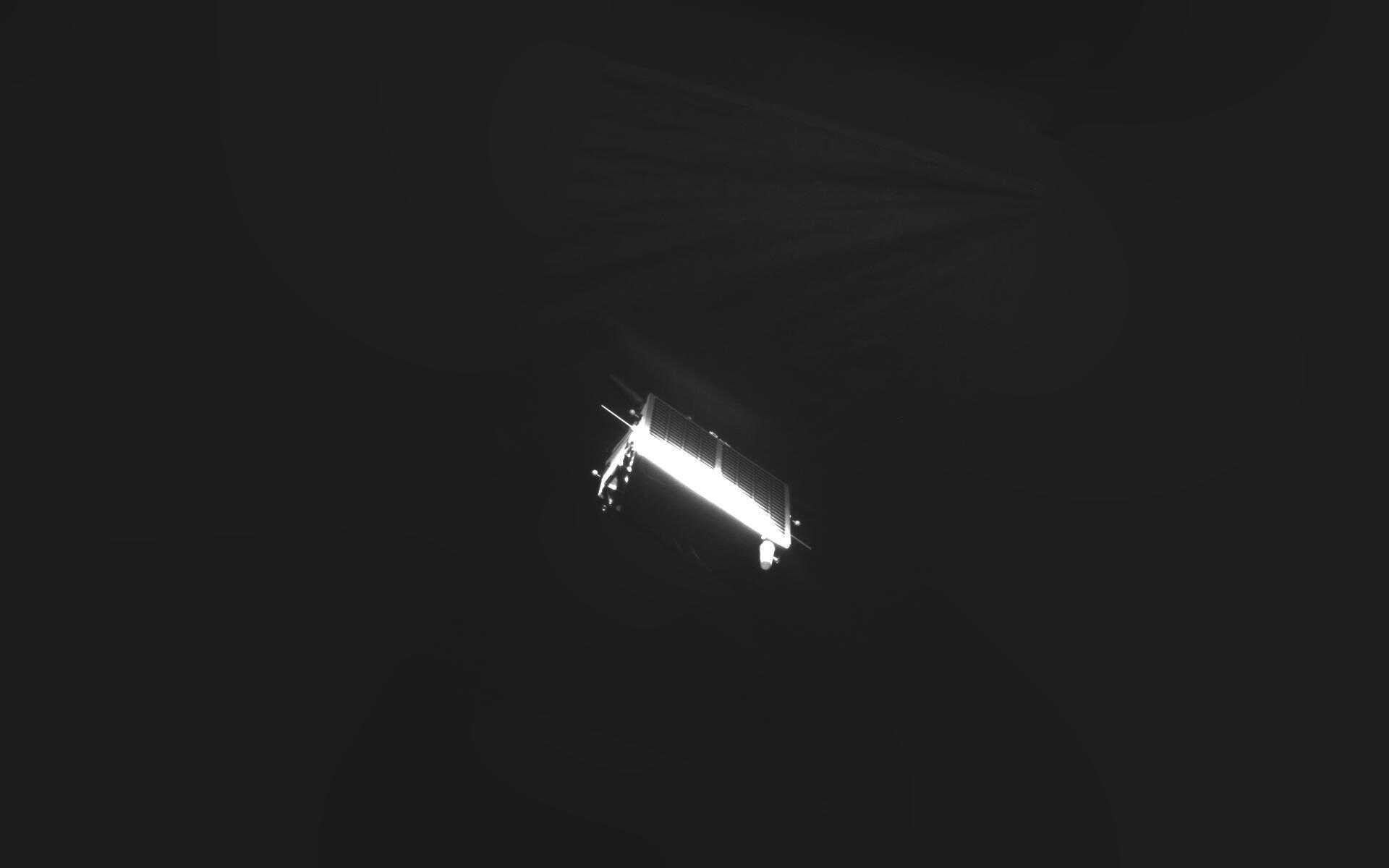} &
\includegraphics[width=\myw,  ,valign=m, keepaspectratio,] {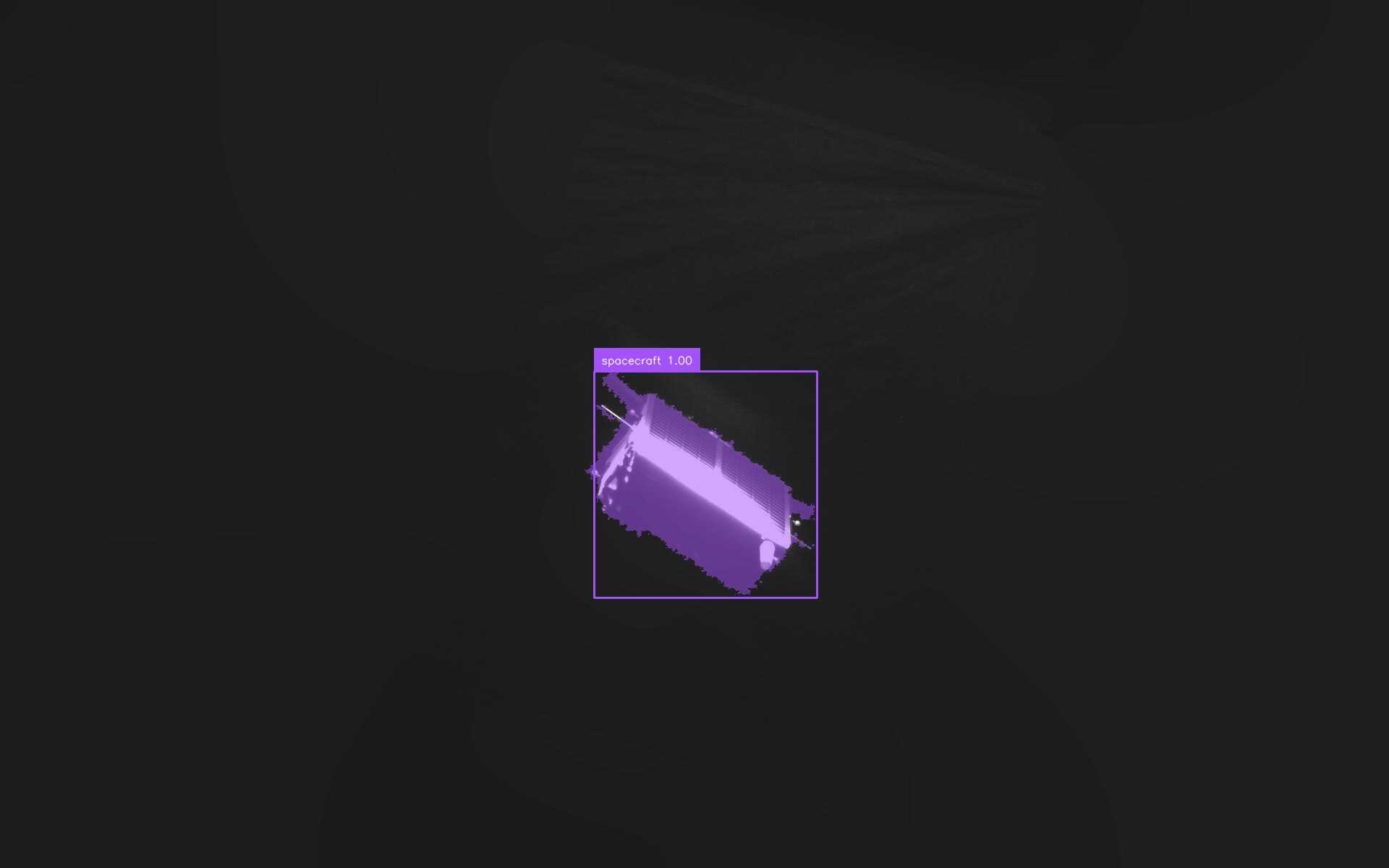} &
\includegraphics[width=\myw,  ,valign=m, keepaspectratio,] {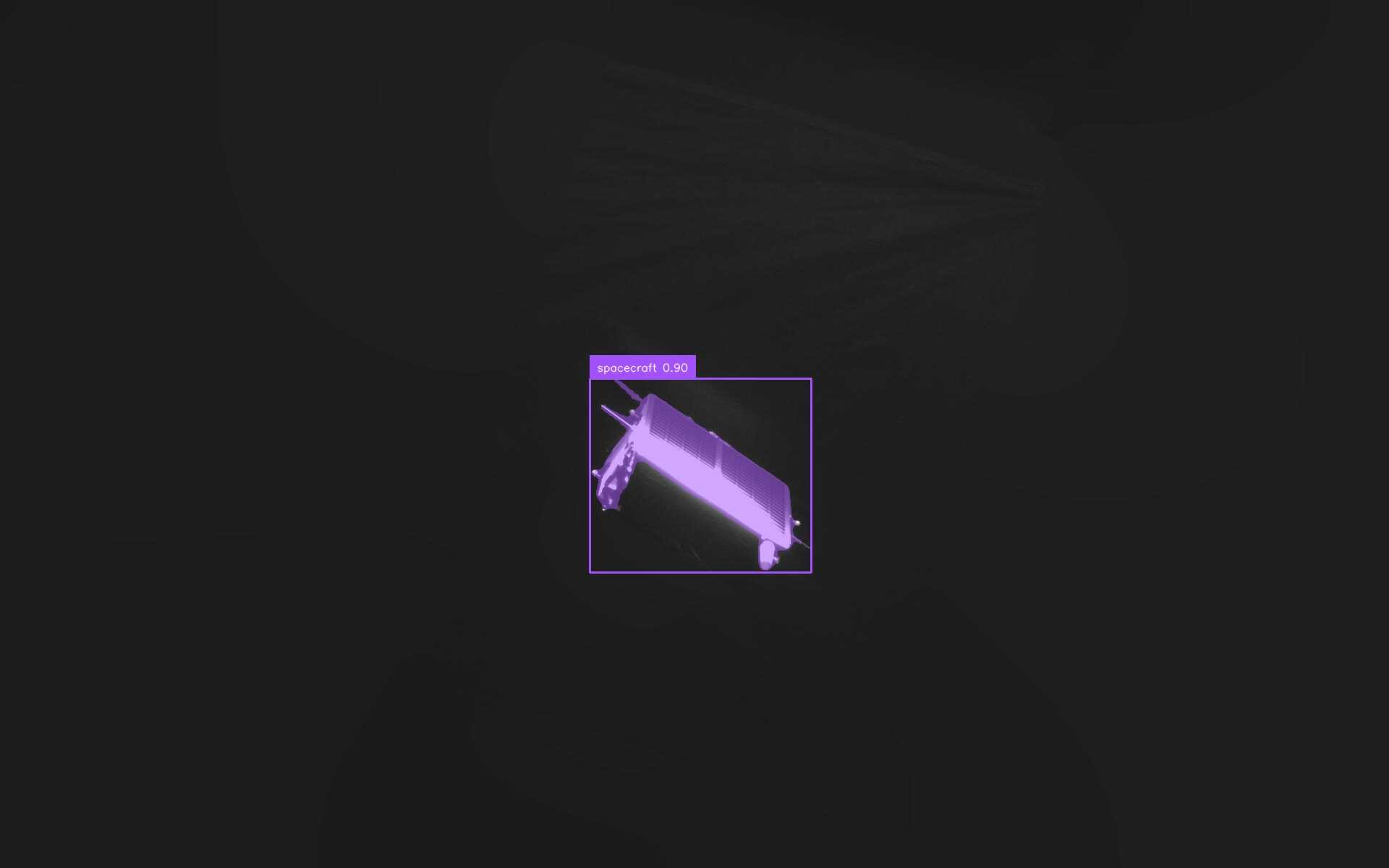} & 
\includegraphics[width=\myw,  ,valign=m, keepaspectratio,] {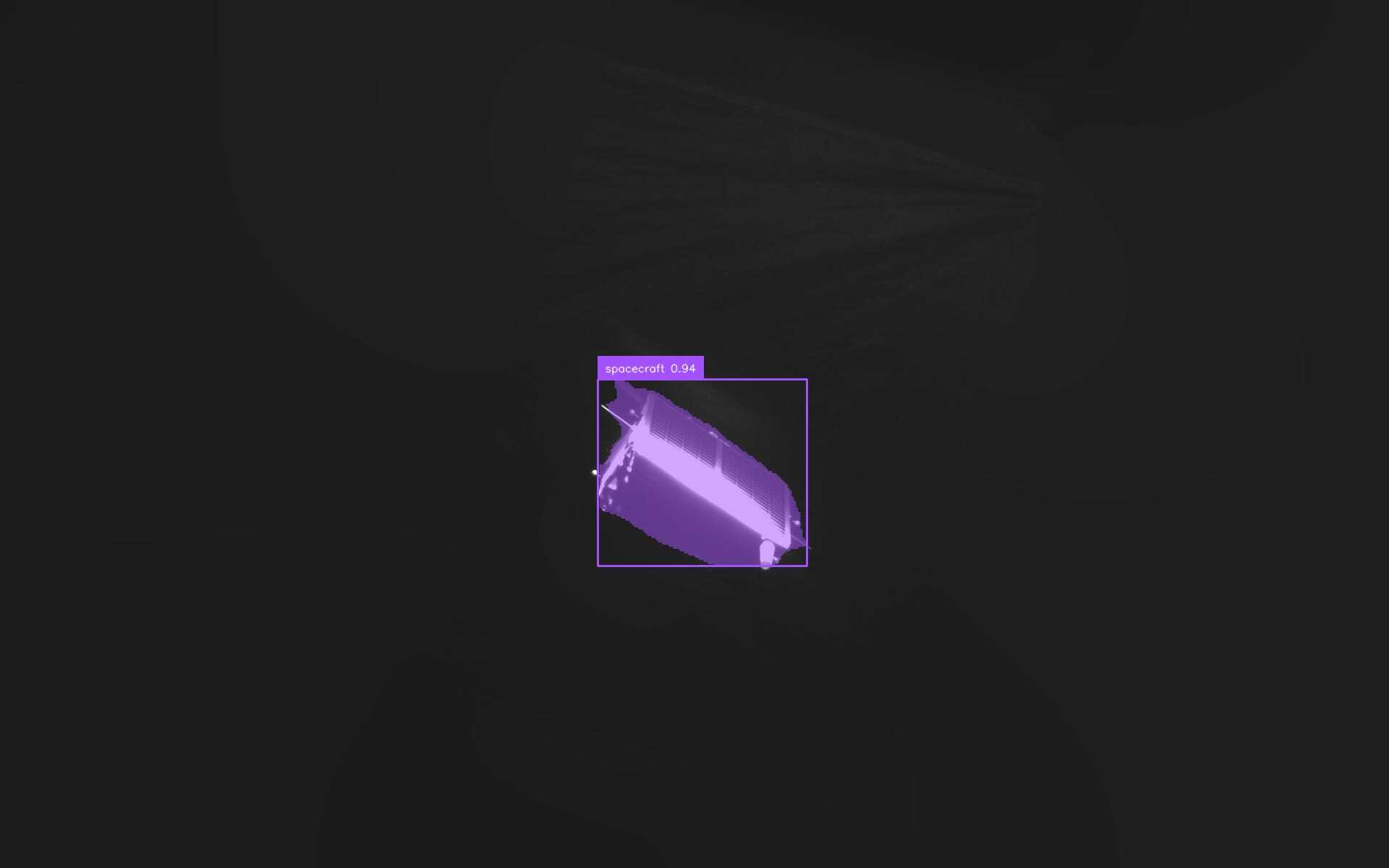}\\
{\rotatebox[origin=t]{90}{\textit{\textbf{SunLamp}}}}  &
\includegraphics[width=\myw,  ,valign=m, keepaspectratio,] {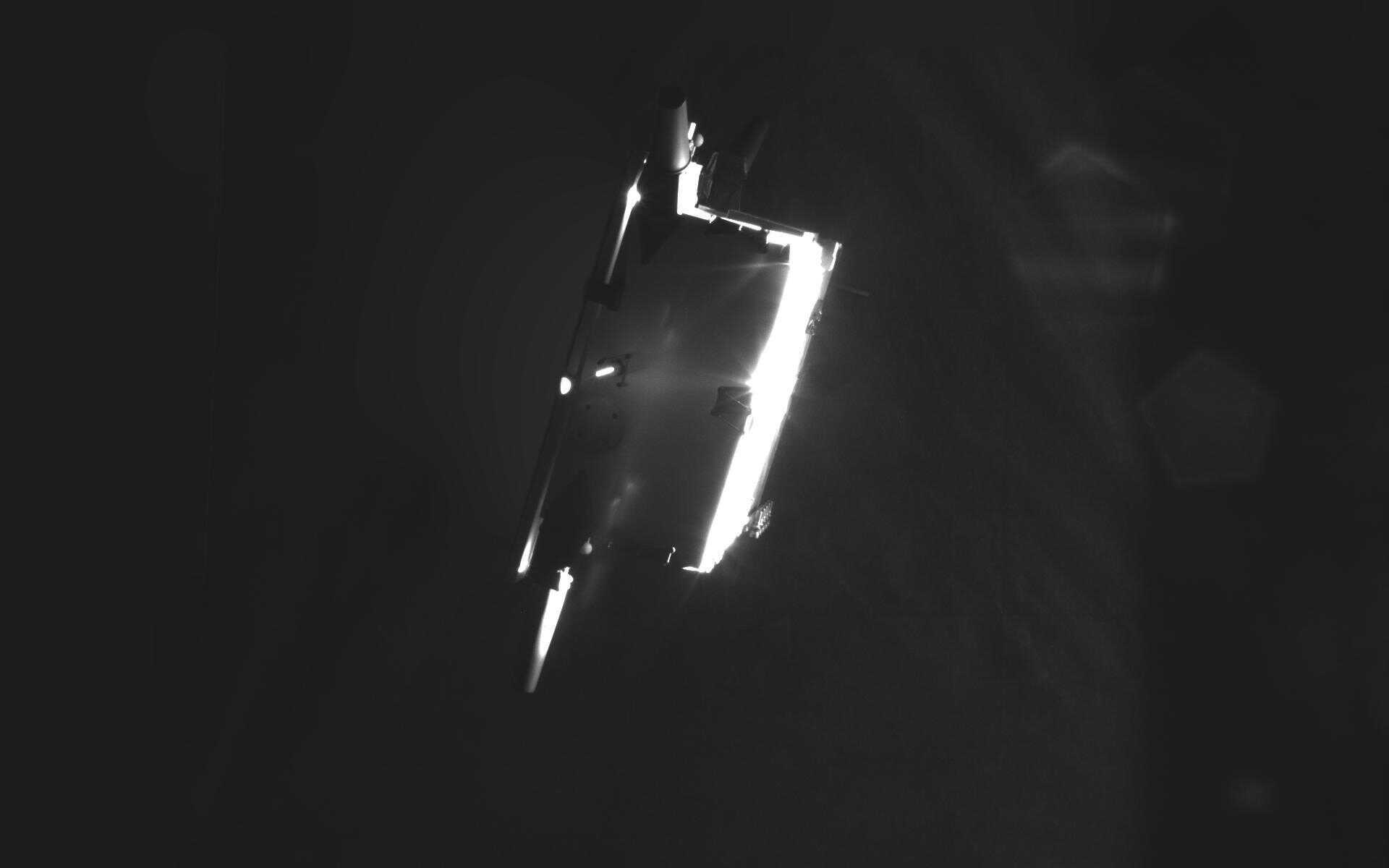} &
\includegraphics[width=\myw,  ,valign=m, keepaspectratio,] {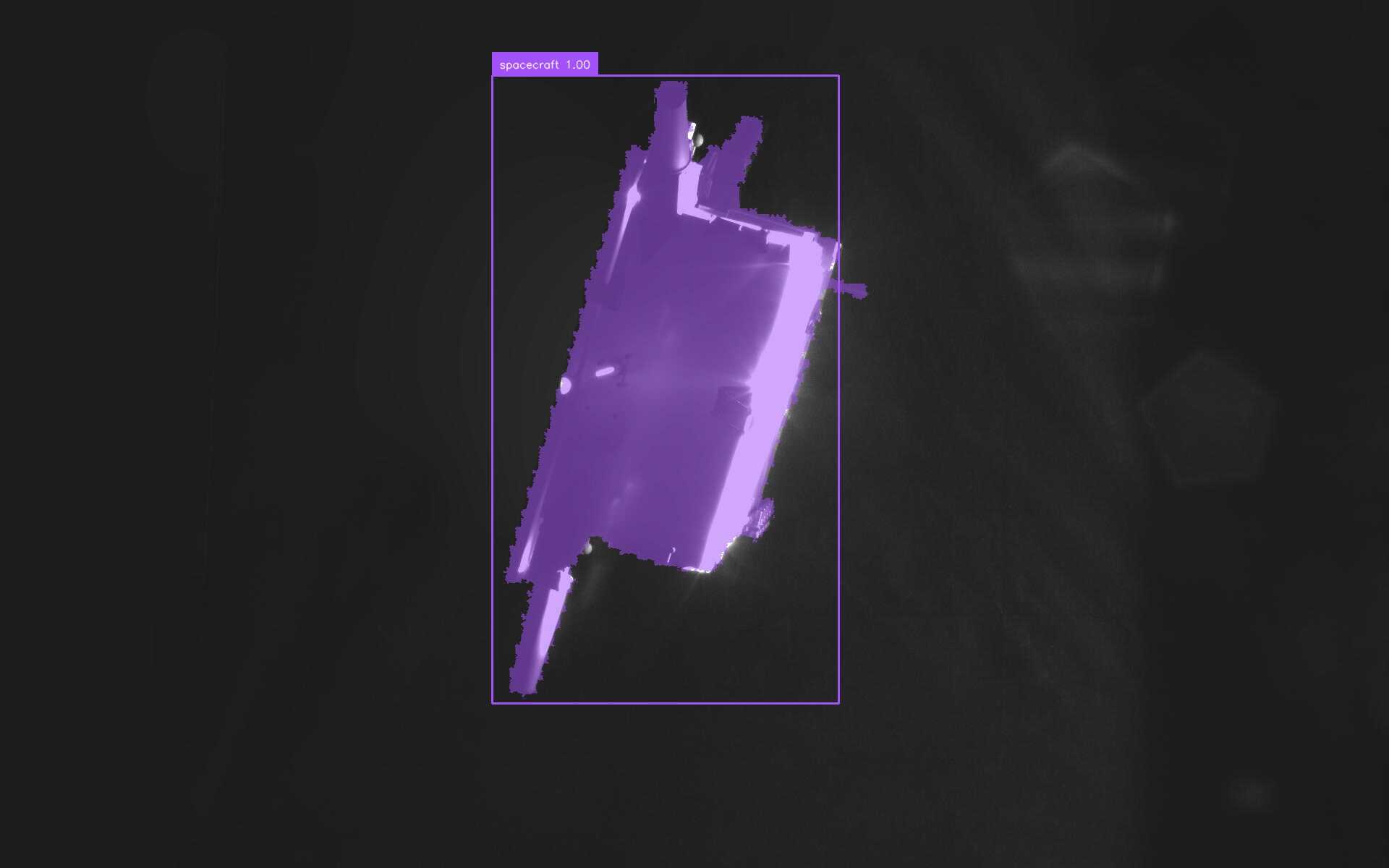} &
\includegraphics[width=\myw,  ,valign=m, keepaspectratio,] {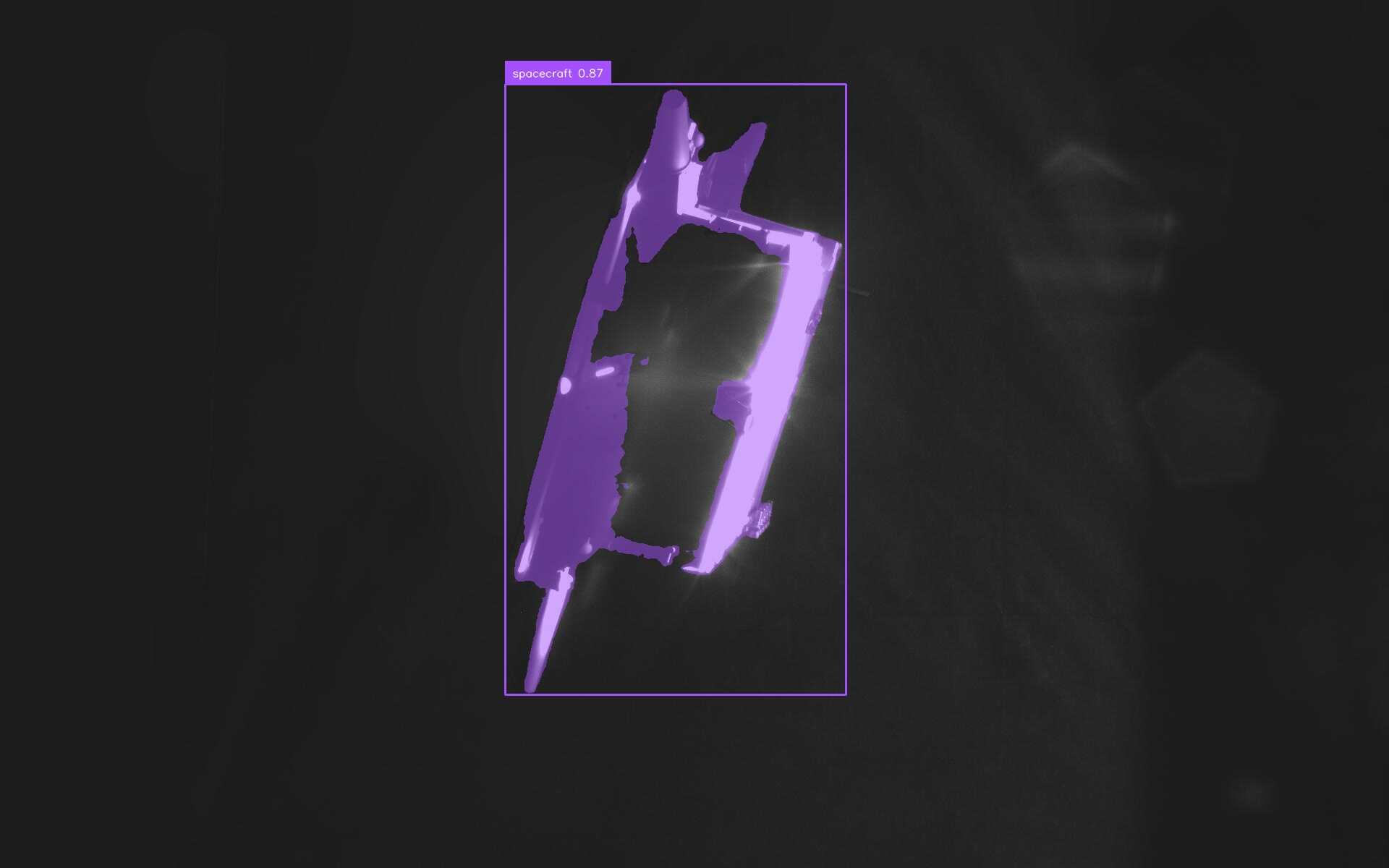} & 
\includegraphics[width=\myw,  ,valign=m, keepaspectratio,] {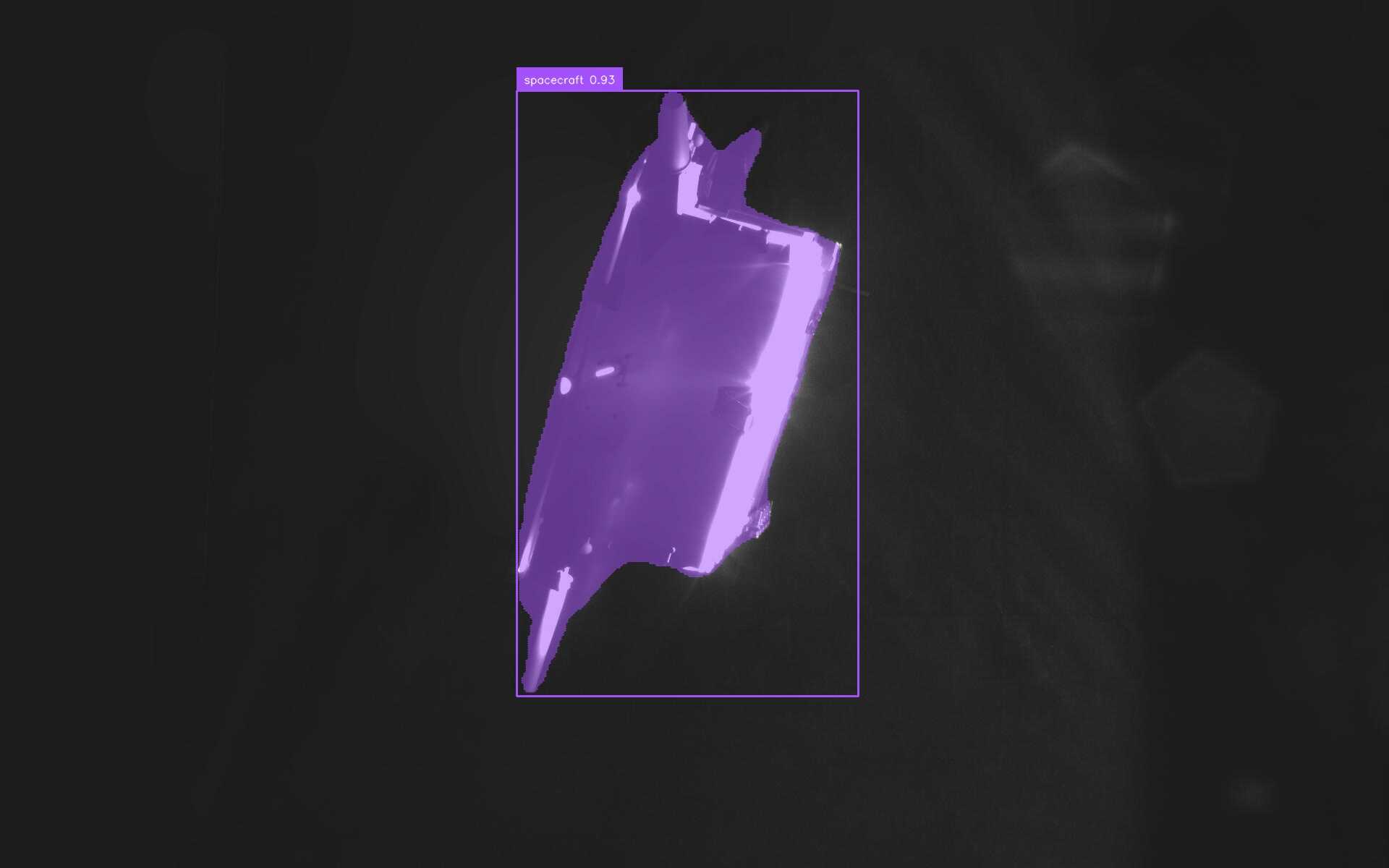}\\

{\rotatebox[origin=t]{90}{\textit{\textbf{TANGO}}}}  &
\includegraphics[width=\myw,  ,valign=m, keepaspectratio,] {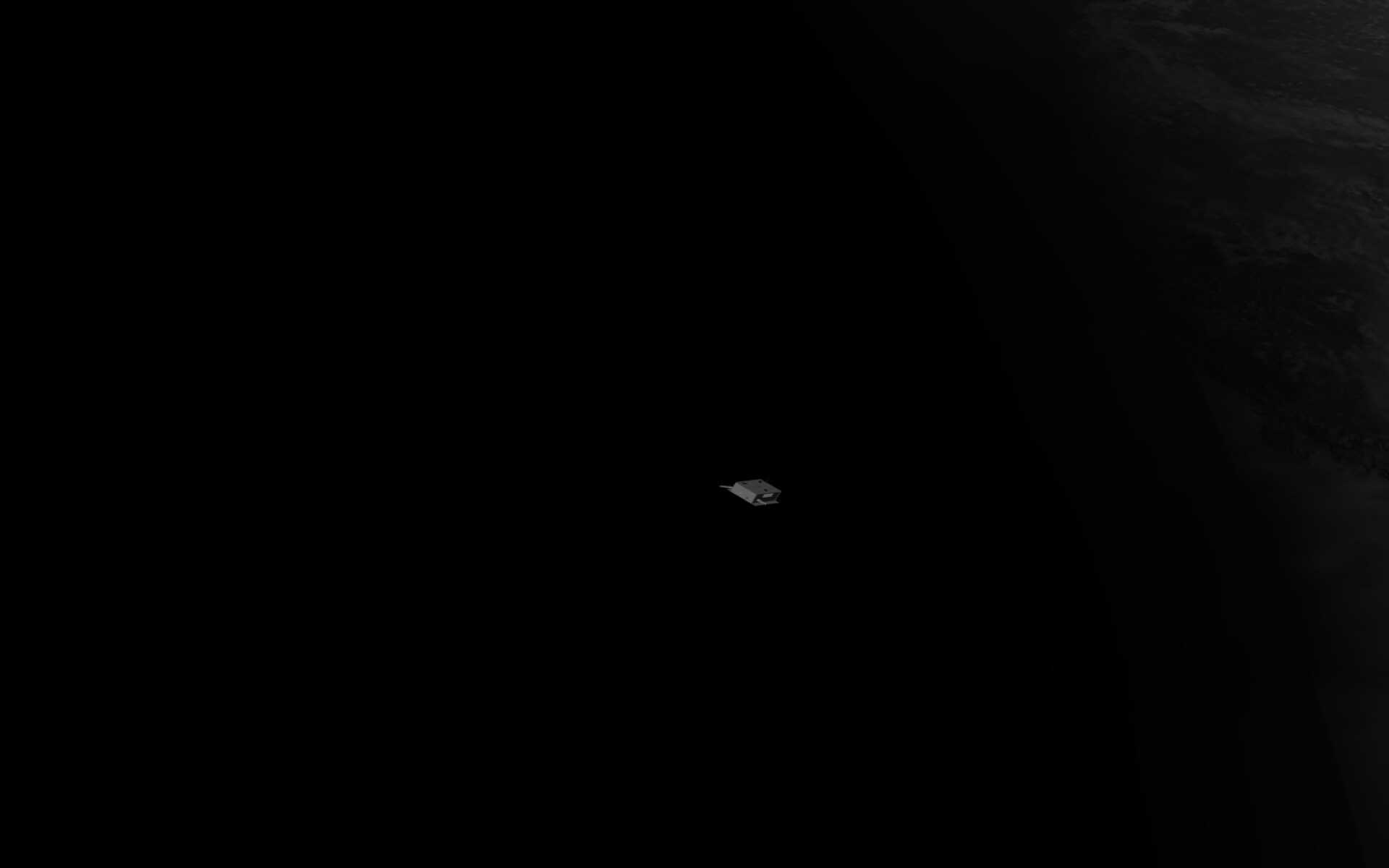} &
\includegraphics[width=\myw,  ,valign=m, keepaspectratio,] {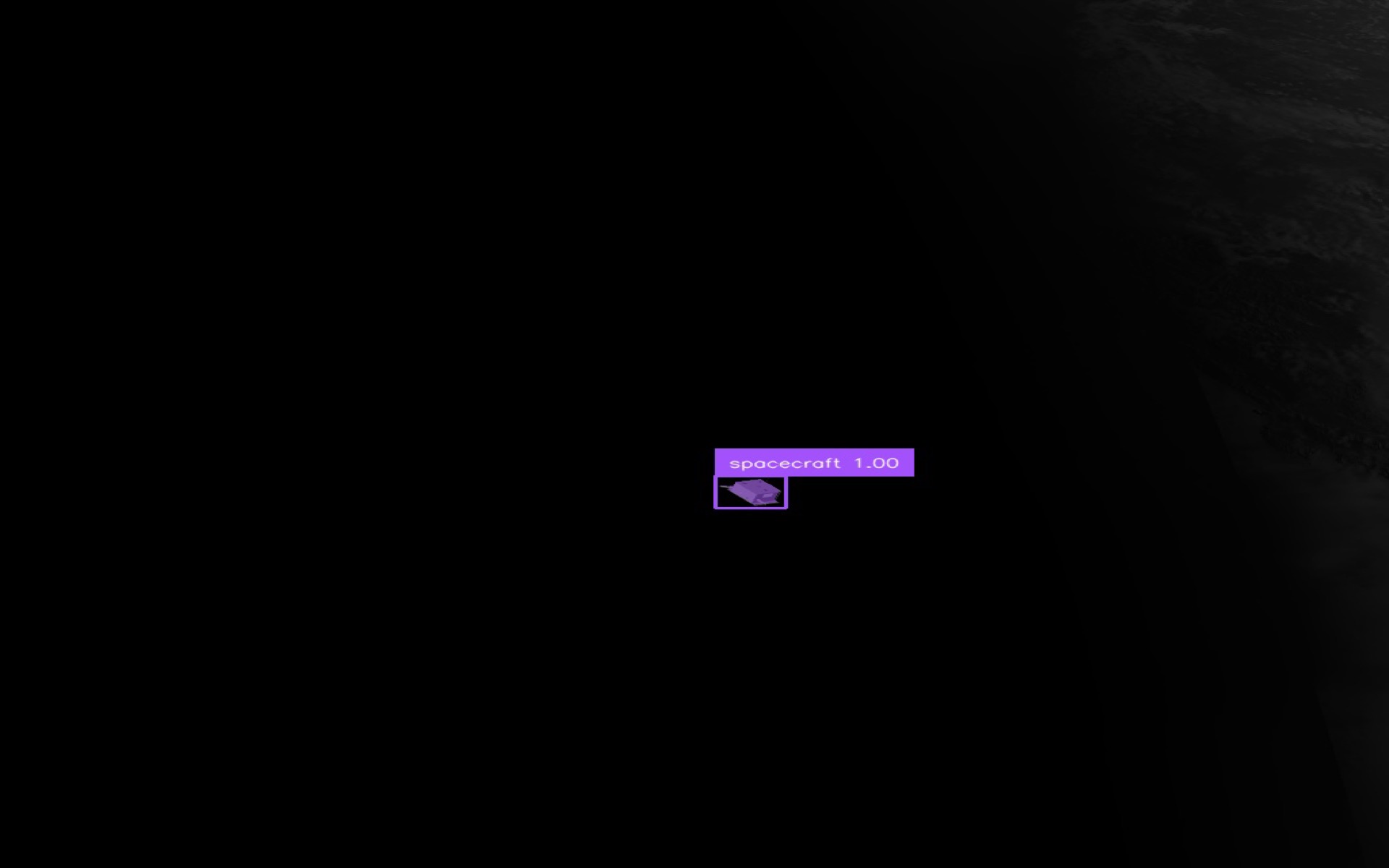} &
\includegraphics[width=\myw,  ,valign=m, keepaspectratio,] {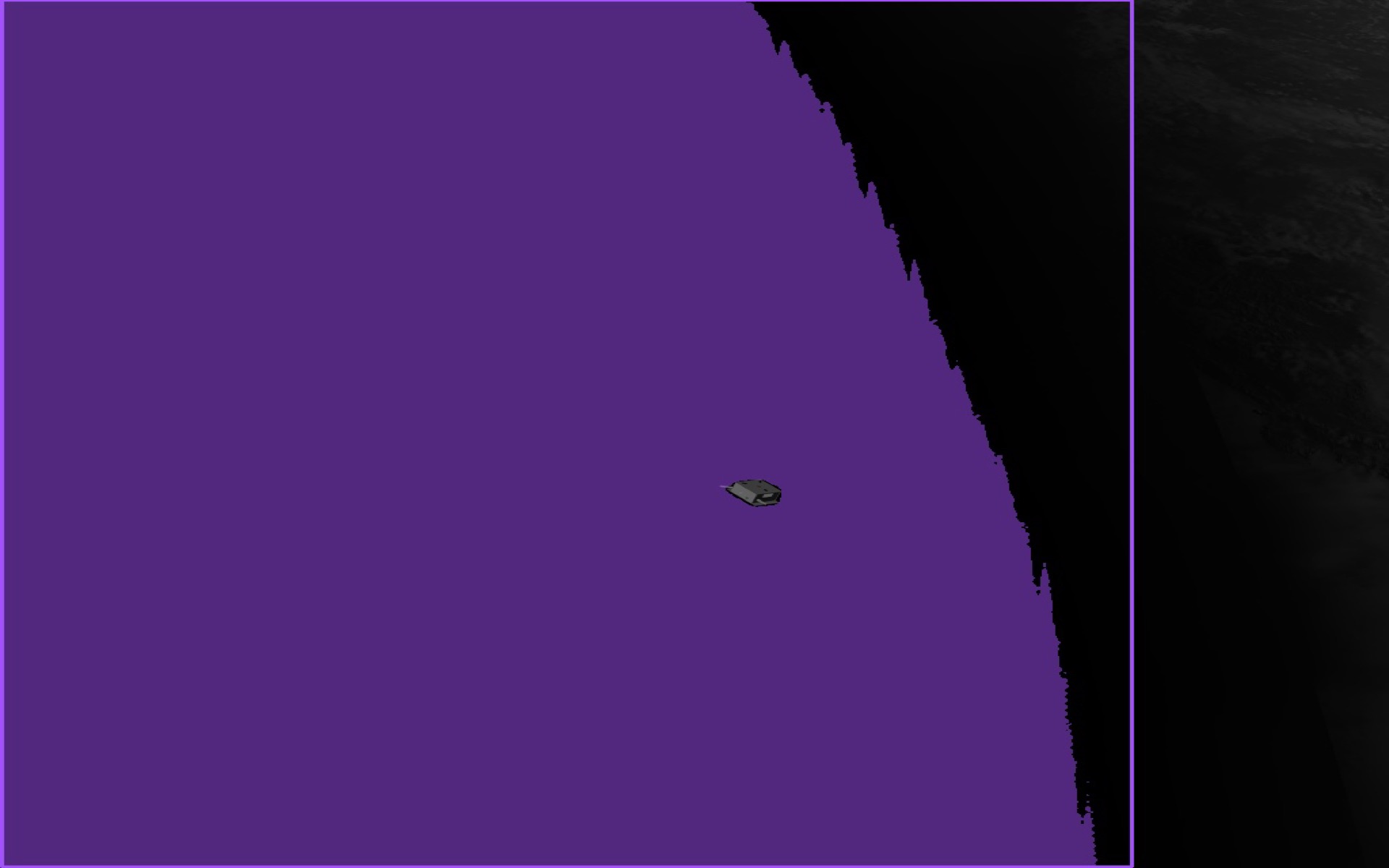} & 
\includegraphics[width=\myw,  ,valign=m, keepaspectratio,] {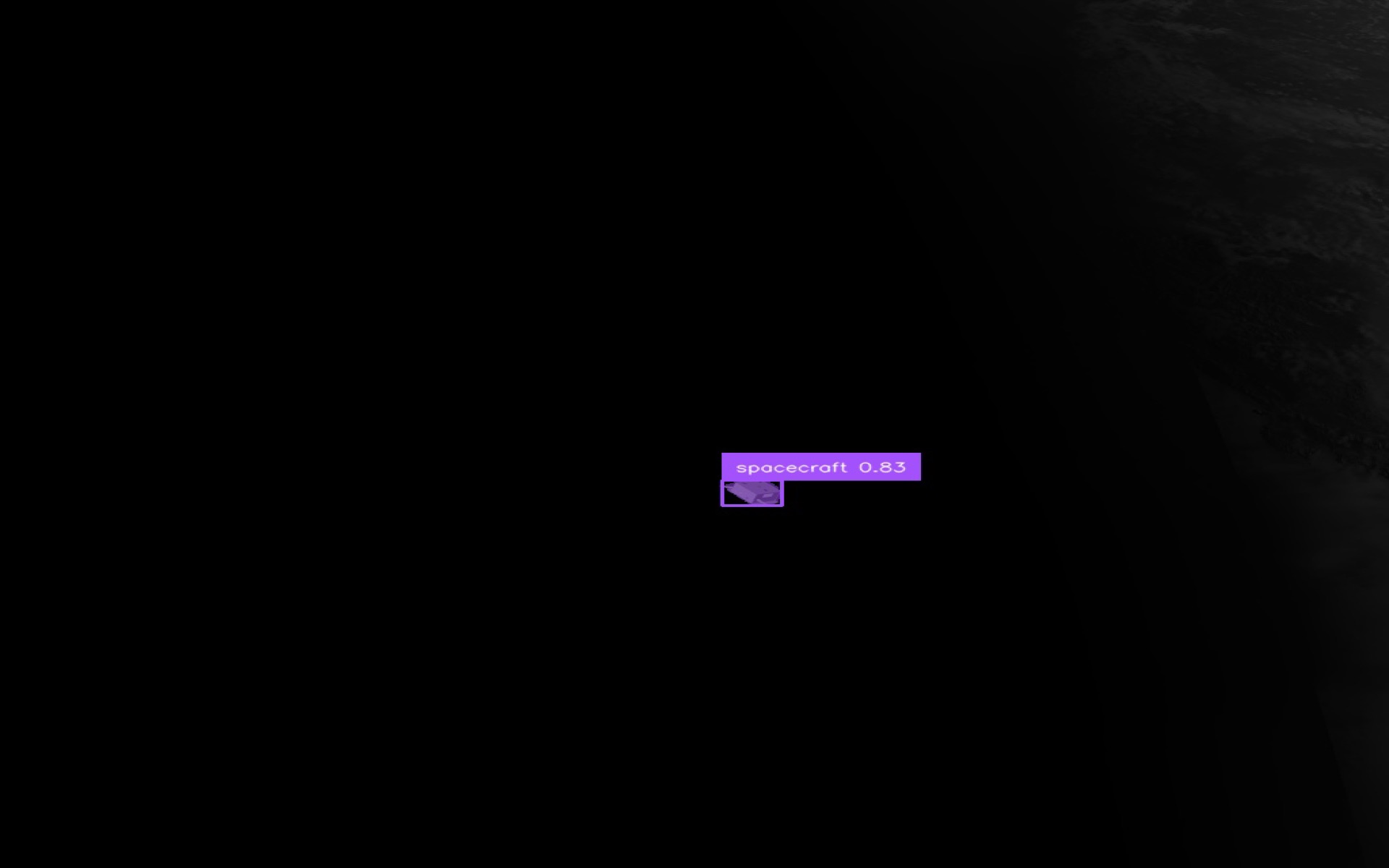}\\

{\rotatebox[origin=t]{90}{\textit{\textbf{TANGO}}}}  &
\includegraphics[width=\myw,  ,valign=m, keepaspectratio,] {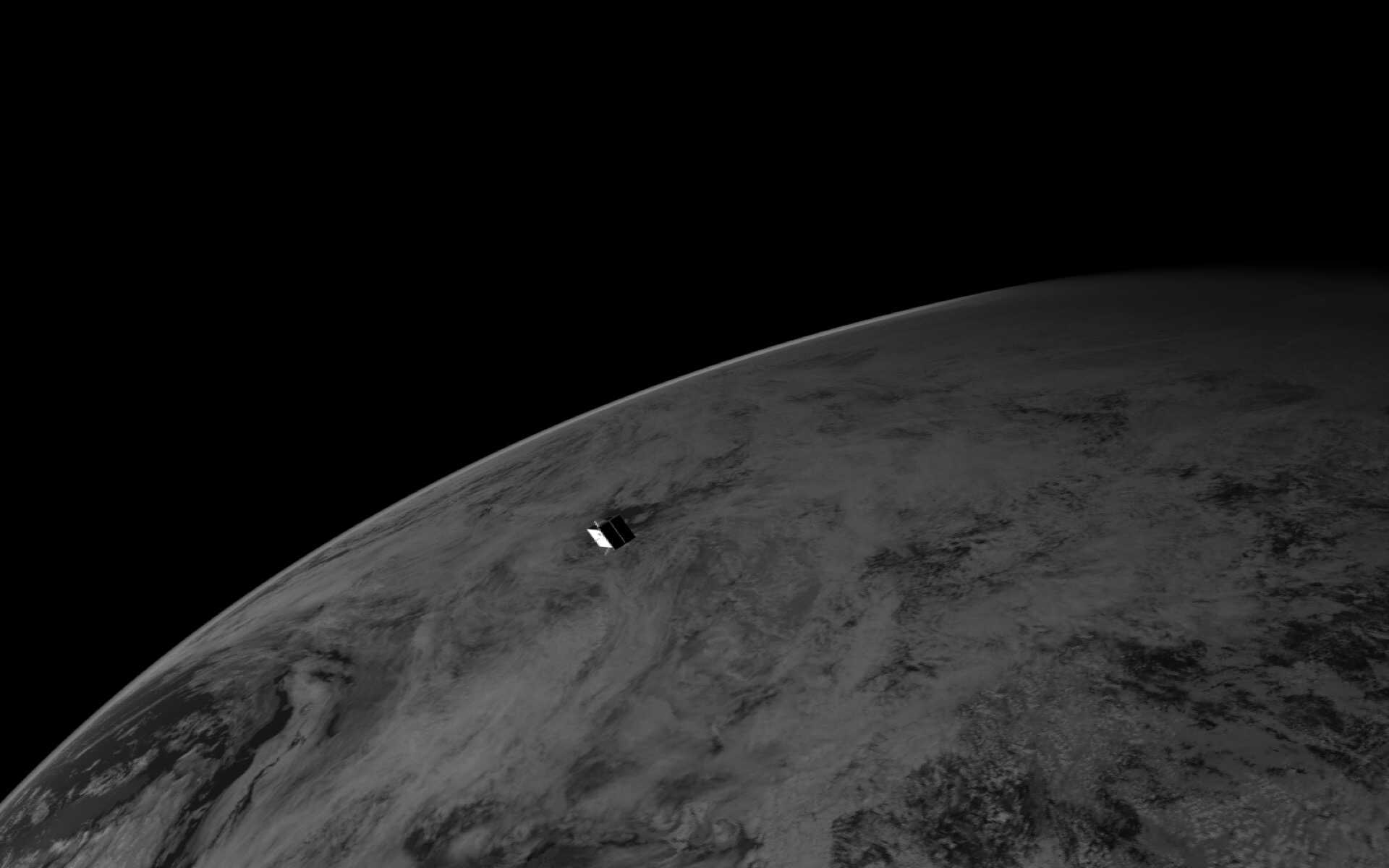} &
\includegraphics[width=\myw,  ,valign=m, keepaspectratio,] {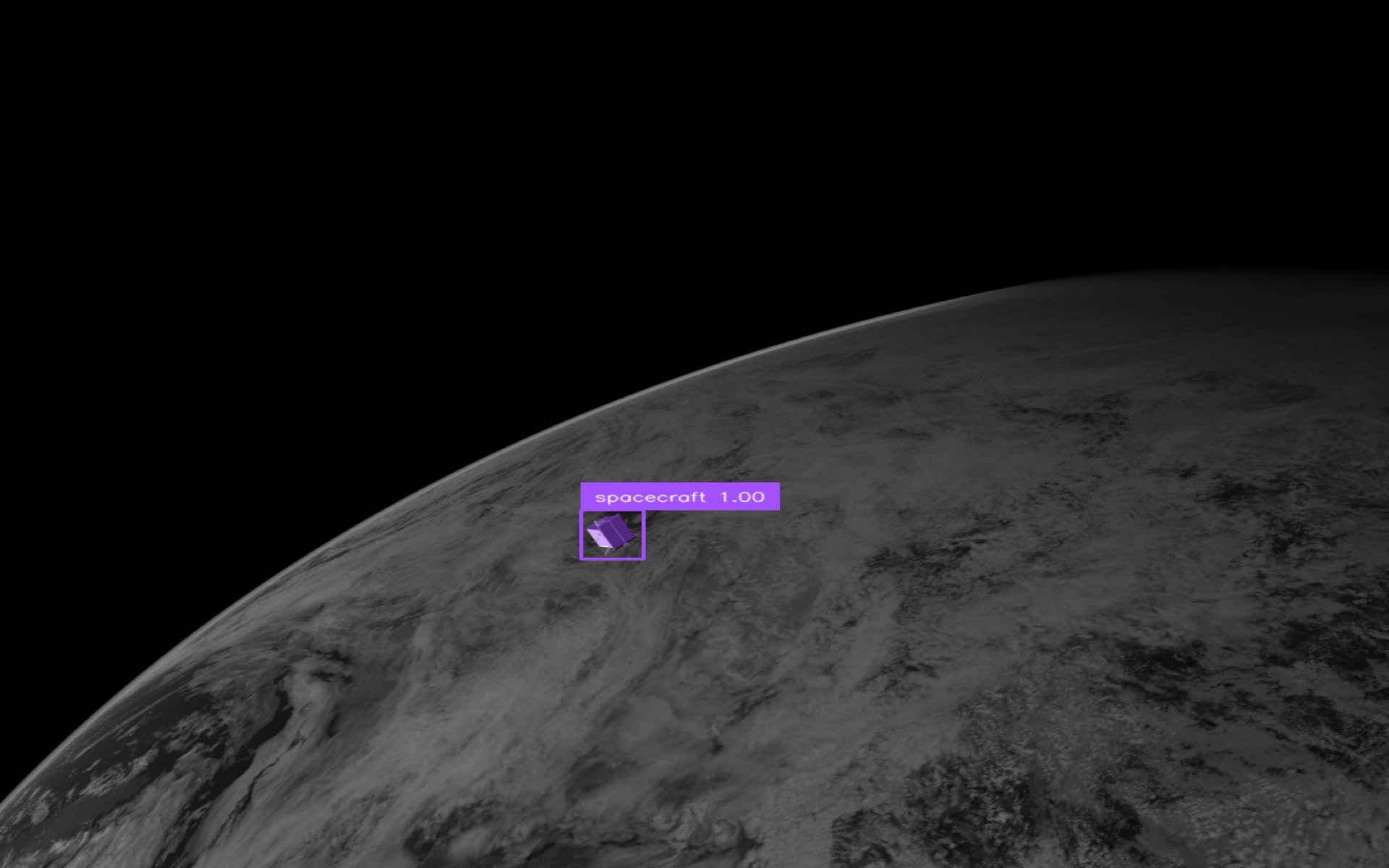} &
\includegraphics[width=\myw,  ,valign=m, keepaspectratio,] {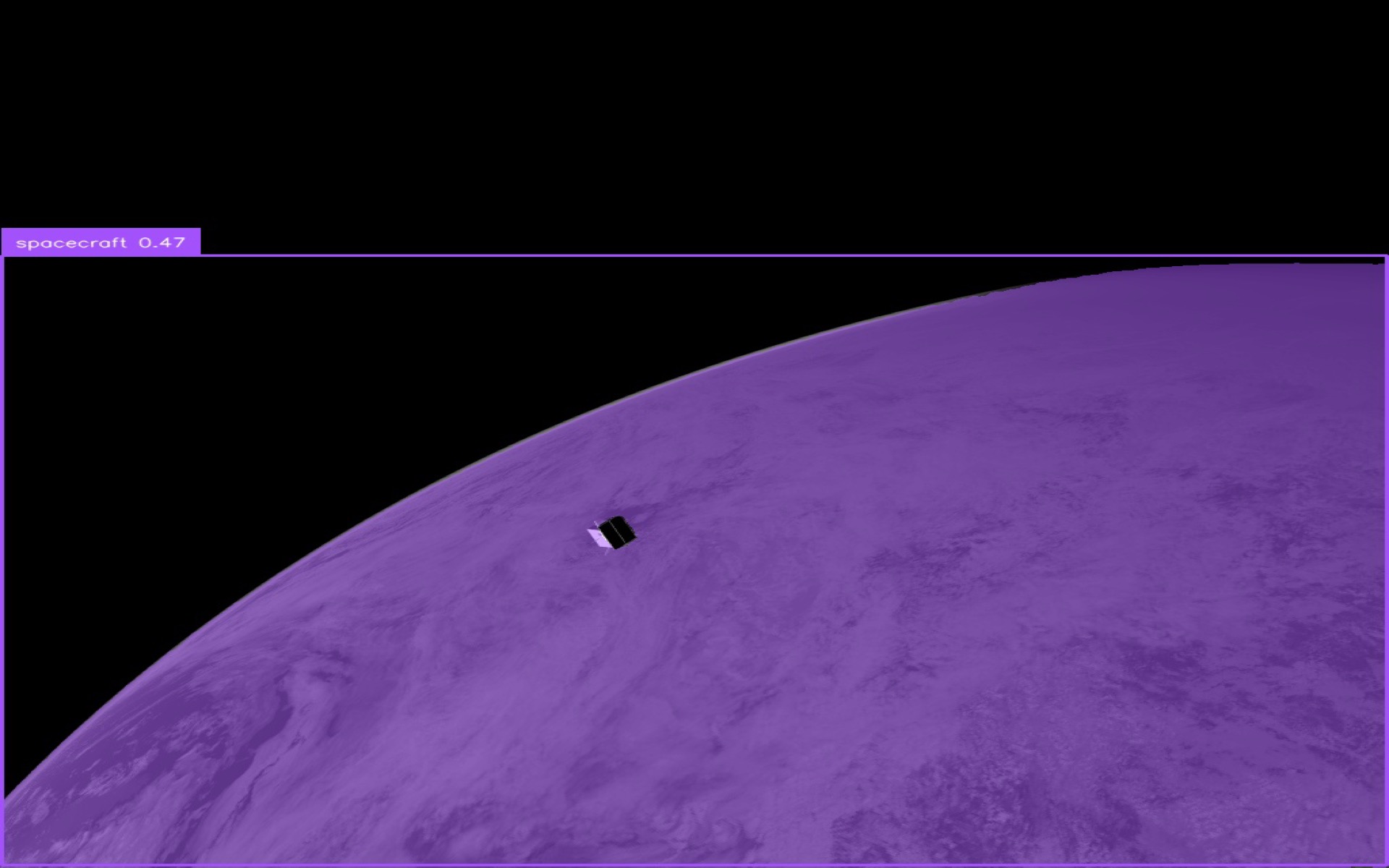} & 
\includegraphics[width=\myw,  ,valign=m, keepaspectratio,] {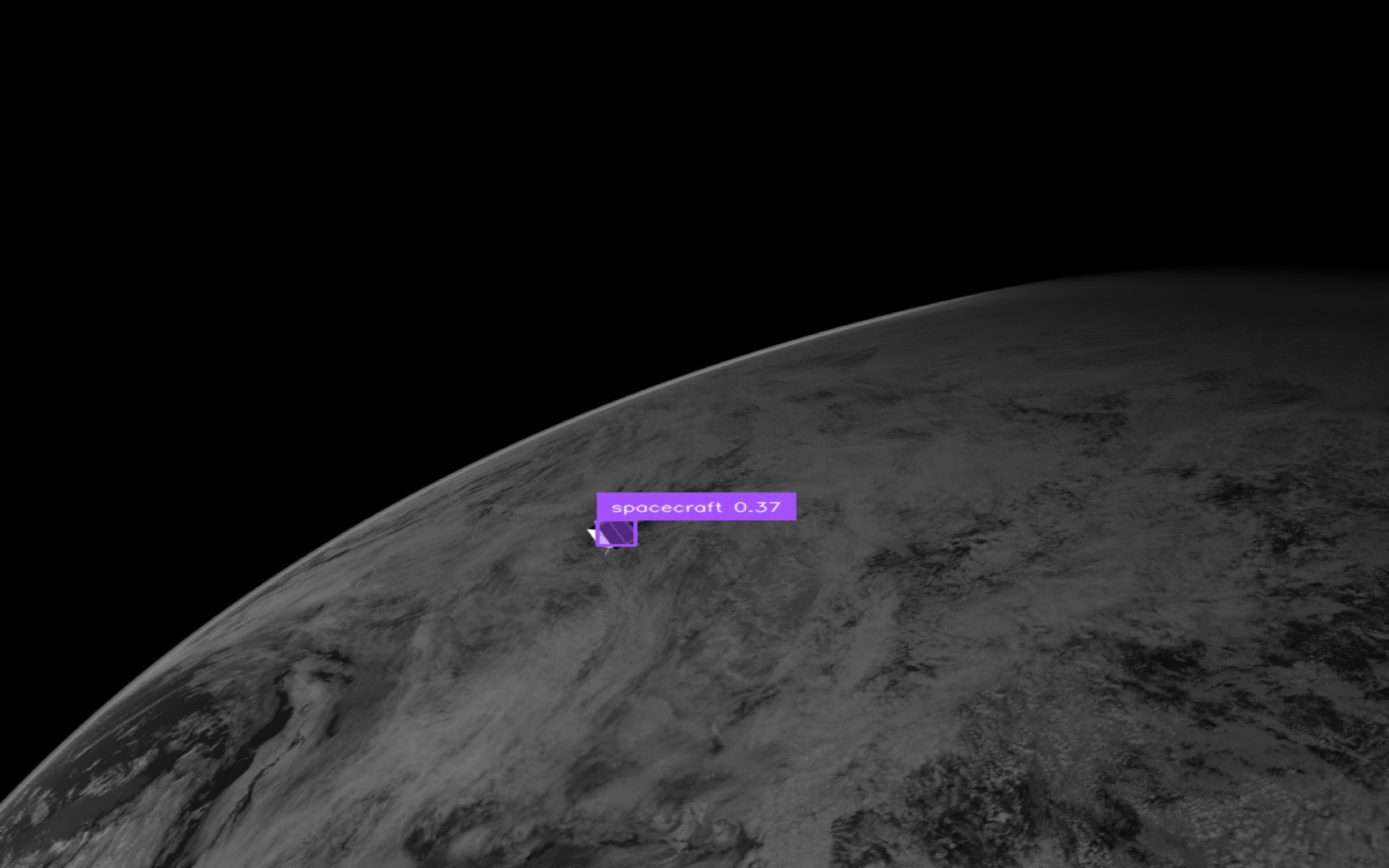}\\

\end{tabular}
\caption{Visual comparison of zero-shot {\VFM} and our student model predictions. Our method greatly improves {\VFM} predictions.}
\label{fig:visuals}
\vspace{-3mm}
\end{figure*}

%% file: sec/5_conclusion.tex
\section{CONCLUSION}

In this work, we present an end-to-end spacecraft detection and segmentation pipeline that does not rely on any form of manually labeled data. Our method exploits the zero-shot capabilities of recent {\VFM}s. It first extracts object bounding boxes and instance masks using a {\VFM}, then utilizes these pseudo-labels in an iterative teacher-student distillation framework. Extensive experiments on three recent large-scale space datasets show that our method significantly improves initial zero-shot {\VFM} results and reduces the gap to fully supervised training.  